\documentclass[preprint,11pt,number]{elsarticle}
\biboptions{sort&compress}
\usepackage[
left=1in,
right=1in,
top=1in,
bottom=1in
]{geometry}

\usepackage{xcolor}
\usepackage{float}
\usepackage{graphicx}
\usepackage{amsmath}
\usepackage{float}
\usepackage{booktabs}
\usepackage{array}
\usepackage{subcaption}
\usepackage{tabularx}
\usepackage{makecell}
\usepackage[normalem]{ulem}
\usepackage[font=footnotesize]{caption}
\usepackage{amssymb}

\usepackage{algorithm} 
\usepackage{algpseudocode}
\usepackage{algorithmicx}

\makeatletter
\def\ps@pprintTitle{%
  \let\@oddhead\@empty%
  \let\@evenhead\@empty%
  \def\@oddfoot{}%
  \let\@evenfoot\@oddfoot}

\begin{document}
\begin{frontmatter}

% \maketitle
\title{AI-Augmented Adaptive Digital Twin Modeling for Brain Tumor Evolution Prediction and Treatment Scheduling}

\author{{Wenxi Liu}}

\author{{Michael Trimboli}}

\author{and Xianqi Li\corref{cor1}}

\cortext[cor1]{Corresponding author}

\affiliation{organization={Department of Mathematics and Systems Engineering},
            addressline={Florida Institute of Technology}, 
            city={Melbourne},
            state={FL},
            postcode={32901}, 
            country={USA}}
\begin{abstract}

\noindent \textbf{Background and Objective:}
Brain tumor evolution is spatially heterogeneous and  shaped by patient-specific treatment responses and the surrounding brain microenvironment. Predicting this evolution is important for treatment scheduling, yet patient-specific tumor dynamics are only partially observed and altered by therapy. We developed an AI-augmented adaptive digital twin (DT) framework that integrates mechanistic tumor modeling, learned model-form correction, patient-specific updating, and constrained treatment scheduling.

\noindent \textbf{Methods:}
We constructed a patient-data-informed synthetic testbed combining imaging-derived anatomy and tumor initialization with controlled treatment-conditioned longitudinal trajectories. The hybrid reaction–diffusion (RD)–residual model retains an interpretable mechanistic backbone, while a 3D residual network learns structured prediction errors not captured by the RD model. During recursive rollout, the residual network is updated using newly available patient-specific observations while the RD backbone remains fixed, reducing accumulated forecasting drift. The updated DT is then coupled with model predictive control (MPC) for constrained chemotherapy and radiotherapy scheduling.

% \textbf{Results:}
% The hybrid RD--residual model reduced masked voxel-wise mean squared error by $84.3\%$ and increased Dice overlap by $43.5\%$ relative to the RD baseline under the dense forecasting setting. DT updating further reduced mean squared error by $45.9\%$ and increased Dice overlap by $9.6\%$ relative to the non-updated hybrid model. MPC-based scheduling reduced final tumor burden by $22.4\%$ relative to a fixed-schedule comparator under the terminal-burden objective.
% \textbf{Results:}
\noindent \textbf{Results:}
The hybrid RD--residual model reduced masked voxel-wise mean squared error by 84.3\% and increased Dice overlap by 43.5\% relative to the RD baseline.
Patient-specific adaptation model reduced mean squared error by 45.9\% and increased Dice by 9.6\%.
MPC-based scheduling achieved a 22.4\% median paired reduction in final tumor burden relative to a fixed schedule under the terminal-burden objective.

\noindent \textbf{Conclusions:}
The proposed framework unifies mechanistic prediction, AI-based model correction, patient-specific adaptation, and treatment-oriented decision making. The study provides controlled patient-data-informed synthetic validation and supports subsequent preclinical and longitudinal clinical evaluation.

\end{abstract}


\end{frontmatter}

\section{Introduction}
% \subsection{Overview}
Glioblastoma and other aggressive brain tumors remain among the most challenging cancers to treat because of their invasive growth, spatial and temporal heterogeneity, and variable treatment response across patients~\cite{stupp2005radiotherapy,ostrom2022cbtrus,louis20212021}. For an individual patient, treatment decisions depend not only on the tumor burden observed at the current clinical assessment, but also on how the disease may evolve under future interventions. Clinical imaging provides the primary window into this process by revealing tumor location, surrounding anatomy, tissue structure, edema, necrosis, and treatment-associated changes~\cite{bakas2022university}. However, longitudinal observations are often sparse, irregular, and confounded by surgery, radiotherapy, chemotherapy, corticosteroid use, pseudoprogression, necrosis, and clinician-selected treatment changes~\cite{wen2010updated}. Consequently, clinical follow-up data rarely provide sufficiently frequent observations under controlled treatment conditions to identify patient-specific tumor dynamics or evaluate how the same patient might respond to alternative treatment schedules.

These limitations motivate the development of digital twin (DT) models for brain tumors. A tumor DT is a computational representation of an individual patient’s disease that is initialized from available data, simulated forward to forecast future evolution, and updated as new observations become available~\cite{fertig2021forecasting,bjornsson2019digital}. Clinical imaging can provide the anatomical context and tumor state needed for initialization and updating, while treatment records characterize interventions that shape subsequent evolution~\cite{wu2022integrating}. In this sense, a tumor DT is more than a static image predictor: it is a dynamic model that links current observations with forward simulation and can be recalibrated as the disease and treatment course evolve.

%\delete{Such a model can also provide a computational basis for evaluating alternative future treatment schedules.}

Mechanistic reaction--diffusion (RD) models provide an interpretable foundation for this purpose. By representing tumor burden as a spatially distributed cellularity-like field, RD models can describe diffusion-driven invasion, local proliferation, anatomical constraints, tissue-dependent migration, carrying-capacity-limited growth, and treatment-induced tumor reduction~\cite{murray2003mathematical,swanson2000quantitative,jbabdi2005simulation}. These properties make RD models a natural bridge between patient-specific anatomical context and forward tumor-evolution simulation~\cite{konukoglu2009image,hormuth2015predicting}. However, they often rely on simplified representations of growth, tissue dependence, treatment response, and local microenvironmental heterogeneity. These approximations can produce systematic errors in tumor density, invasive margins, regional growth patterns, and treatment response. During long-horizon recursive forecasting, such errors may accumulate and substantially distort the predicted tumor burden and morphology.
% Practical RD models, however, must approximate complex tumor biology using a tractable set of parameters and equations. 
% \add{Reaction--diffusion (RD) models describe tumor invasion, proliferation, and treatment response within patient-specific anatomy~\cite{murray2003mathematical,swanson2000quantitative,jbabdi2005simulation,konukoglu2009image,hormuth2015predicting}. Their simplified representations of spatial heterogeneity can produce systematic errors in tumor cellularity and extent. These errors may accumulate during recursive forecasting.}
%  \delete{Brain tumor evolution prediction, however, is not merely an image-to-image prediction problem.}

Learning-based models offer complementary flexibility by representing spatial and temporal patterns that are difficult to specify explicitly in mechanistic equations~\cite{raissi2019physics}. A useful tumor DT must support interpretable forward simulation, recursive forecasting, patient-specific adaptation, uncertainty analysis, and treatment evaluation. Purely data-driven models may approximate observed transitions, but they do not necessarily preserve tumor-growth mechanisms, anatomical constraints, or treatment effects. Hybrid mechanistic–learning models provide a compromise between these two approaches: the mechanistic component supplies explicit dynamical structure, while the learned component corrects systematic behavior not captured by the prescribed equations~\cite{mascheroni2021improving}. In the present framework, the RD model serves as the tumor-evolution backbone, and a 3D residual network learns structured voxel-level prediction errors associated with unresolved spatial heterogeneity and treatment-conditioned evolution.

A further challenge arises during recursive deployment. A model trained for one-step prediction may perform well when conditioned on reference states, but long-horizon prediction requires repeatedly feeding predicted states back into the model. This creates exposure bias and rollout distribution shift~\cite{bengio2015scheduled}. Small local errors can accumulate over time, shifting the predicted trajectory away from the patient-specific evolution and degrading both tumor-burden estimates and spatial morphology. To remain aligned with the evolving patient state, the DT must be updated when new patient-specific observations become available. In the proposed framework, adaptive DT updating is performed by recalibrating the learned residual component during recursive rollout while preserving the mechanistic RD backbone~\cite{wang2022continual}. This design allows the model to retain interpretable tumor-growth structure while adapting the residual correction to the current patient-specific trajectory.

Adaptive forecasting is also important for treatment scheduling. Treatment decisions depend on how the tumor is expected to evolve under candidate future actions; therefore, the DT must function as a forward simulation model for comparing treatment policies. Dynamical tumor models have long been used in optimal-control and adaptive-therapy studies to investigate treatment timing and dose selection~\cite{martin1992optimal}. Model predictive control (MPC) is particularly suitable for this setting because it evaluates candidate treatment sequences over a finite planning horizon, applies the first selected action, updates the system state, and replans as new information becomes available~\cite{mayne2000survey,rawlings2017model,hirata2014model}. This receding-horizon structure naturally complements an adaptive DT: the tumor forecast is revised as the simulated state evolves, and treatment actions are selected using the updated model. In this study, the DT is coupled with MPC to evaluate chemotherapy and radiotherapy action policies under timing, dose, toxicity, and bounded-uncertainty constraints. 
\begin{figure}[t]
\centering
\includegraphics[width=\textwidth]{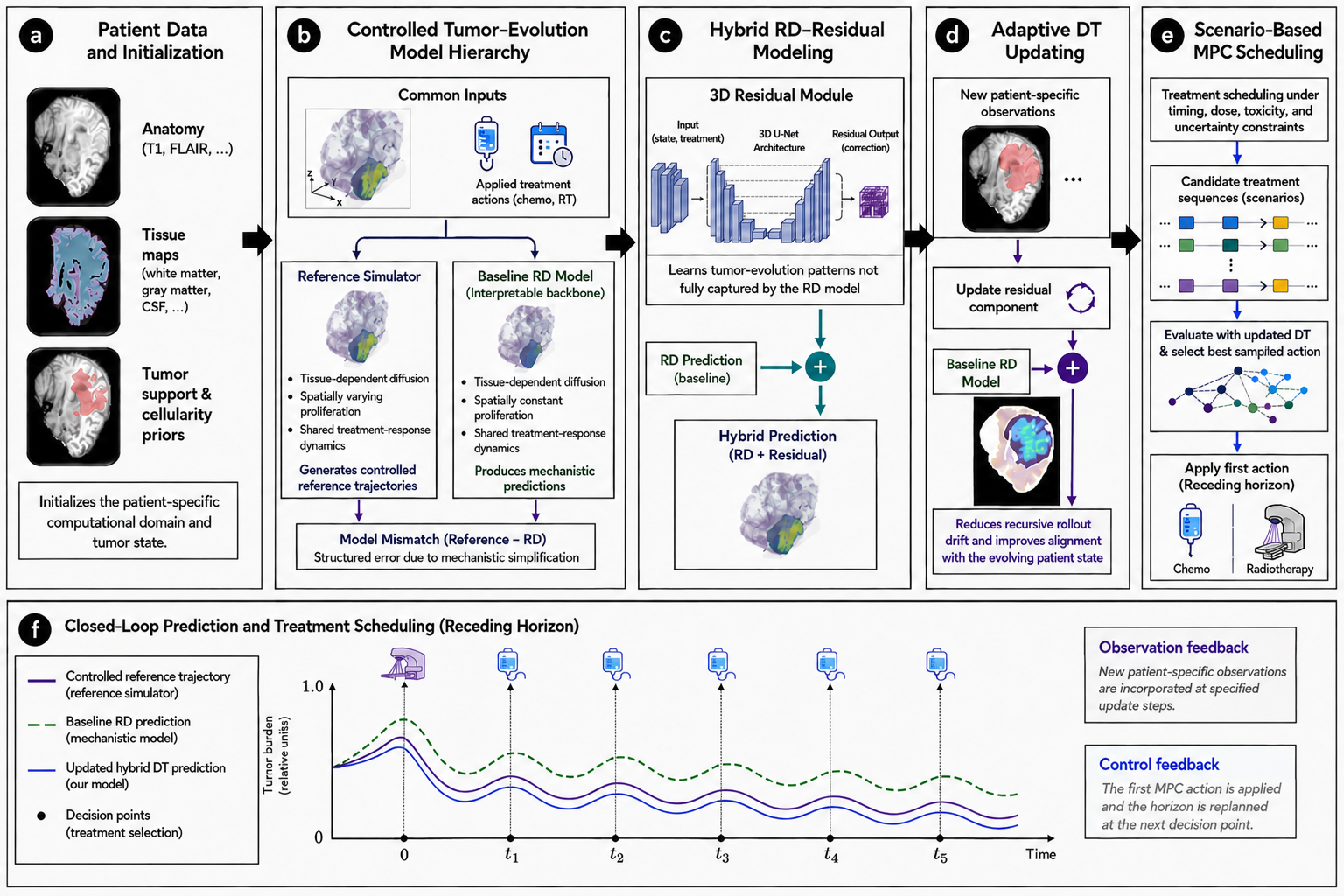}
\caption{
Overview of the proposed AI-augmented adaptive DT framework for brain tumor evolution prediction and treatment-scheduling simulation. Patient imaging data define the computational domain, tissue maps, tumor support, and initial tumor state. A controlled model hierarchy generates reference trajectories and baseline RD predictions, whose discrepancy defines structured model-form mismatch. A 3D residual module corrects the baseline RD prediction, and newly available observations are used to update the residual component during recursive rollout. The updated DT is then coupled with scenario-based MPC to evaluate feasible chemotherapy and radiotherapy schedules under timing, dose, toxicity, and uncertainty constraints. The generated schedules are used only for computational evaluation.
}
\label{fig:online_adaptive_digital_twin_overview}
\end{figure}
% The resulting schedules are used only to evaluate the computational framework and are not intended as clinical treatment recommendations.
% \add{Model predictive control (MPC) uses updated forecasts to select treatment actions over a receding horizon~\cite{mayne2000survey,rawlings2017model,hirata2014model}. We couple the adapted DT with MPC to evaluate chemotherapy and radiotherapy schedules under timing, dose, toxicity, and bounded state-uncertainty constraints.}

% In this work, we propose an AI-augmented adaptive DT framework for brain tumor evolution prediction and treatment scheduling. The overall workflow is illustrated in Fig.~\ref{fig:online_adaptive_digital_twin_overview}. Patient imaging defines the anatomy and initial tumor state, while a controlled simulator generates treatment-conditioned trajectories for evaluation. The hybrid model combines an RD backbone with a 3D residual network that corrects structured prediction errors. The residual network is adapted as new tumor-state observations become available during recursive forecasting. The adapted DT is then coupled with MPC to evaluate feasible chemotherapy and radiotherapy schedules. Recorded treatment actions serve as inputs during forecasting, whereas candidate future actions become decision variables during scheduling. The main contributions of this work are:
Figure~\ref{fig:online_adaptive_digital_twin_overview} summarizes the framework.  The main contributions of this work are:
\begin{itemize}
\item We construct a patient-data-informed synthetic testbed that combines imaging-derived anatomy and tumor initialization with controlled, treatment-conditioned longitudinal trajectories, enabling systematic evaluation of forecasting accuracy, recursive stability, patient-specific adaptation, and treatment scheduling.

\item We formulate a hybrid RD–residual model in which the RD component retains explicit tumor-growth dynamics, while a 3D residual network corrects structured voxel-level errors not captured by the mechanistic model.

\item We develop a rollout-based adaptation strategy that updates the residual network using recent patient-specific observations while preserving the RD backbone.

\item We couple the adapted DT with scenario-based robust MPC to evaluate chemotherapy and radiotherapy action policies under timing, dose, toxicity, and uncertainty constraints.
\end{itemize}

% \delete{Experiments on patient-informed synthetic trajectories show that the RD model captures coarse tumor location and broad temporal behavior but accumulates systematic errors during long recursive rollouts. Residual correction substantially improves reference-conditioned tumor-state prediction, while patient-specific adaptation further improves recursive forecasting under frequent synthetic observations. When embedded in MPC, the adapted DT supports constrained treatment-scheduling simulation and reduces terminal tumor burden relative to a fixed schedule under the selected terminal-burden objective. Together, these results provide a controlled proof of concept for integrating patient-specific initialization, hybrid tumor forecasting, adaptive updating, and treatment-oriented decision making.}

The remainder of this paper is organized as follows. Section 2 reviews related work on mechanistic tumor modeling, tumor DTs, hybrid mechanistic–learning models, adaptive updating, and MPC-based treatment scheduling. Section 3 presents the proposed framework. Section 4 describes the experimental setting, implementation details, forecasting and adaptation protocols, treatment-scheduling procedure, and evaluation metrics. Section 5 reports the forecasting, adaptation, scheduling, and sensitivity-analysis results. Section 6 discusses the findings and limitations, and Section 7 concludes the paper.

\section{Background and Related Work}

\subsection{Mechanistic Tumor Modeling and Patient-Data-Informed DTs}

RD models have been widely used to describe glioma growth as the spatiotemporal evolution of a spatial tumor-state field~\cite{murray2003mathematical,harpold2007evolution}. Early proliferation–invasion models linked tumor growth patterns to net proliferation and migration~\cite{swanson2000quantitative,mandonnet2003continuous}, while subsequent studies incorporated anatomical constraints, tissue-dependent migration, and diffusion imaging to better represent tumor spread through brain tissue~\cite{jbabdi2005simulation,maier2010diffusion,clatz2005realistic}. Previous studies have estimated RD parameters from imaging while accounting for mass effect~\cite{hogea2008image}, incorporated anatomical and diffusion information into patient-specific glioma growth models~\cite{konukoglu2009image,konukoglu2010personalization}, constrained tumor predictions using quantitative magnetic resonance imaging (MRI) \cite{hormuth2015predicting}, and related proliferation--invasion kinetics to radiation response or molecular characteristics~\cite{rockne2010predicting,swanson2008mathematical,baldock2014invasion}. 
% \delete{These studies show that anatomical structure, tissue maps, tumor extent, quantitative imaging, and treatment information can support patient-specific initialization, model calibration, and forward simulation.}

This modeling paradigm is closely aligned with the concept of a tumor DT: an individualized computational model that can be initialized from patient data, simulated forward, and updated as new observations become available~\cite{fertig2021forecasting,bjornsson2019digital}. Recent work in oncology has increasingly explored the integration of imaging and mechanism-based modeling for individualized prediction and treatment-response analysis~\cite{wu2022integrating,wu2025mri,yankeelov2013clinically,chaudhuri2023predictive}. However, tractable mechanistic models necessarily simplify the biological processes that govern tumor progression and treatment response. In addition, longitudinal clinical observations are often sparse, irregular, and treatment-confounded, which limits direct evaluation of patient-specific dynamics and alternative treatment trajectories. These challenges motivate controlled patient-data-informed settings in which individualized anatomy and tumor initialization are retained while known reference trajectories are used for systematic methodological evaluation.
% \add{Oncology DT studies combine imaging and mechanistic modeling for individualized prediction and treatment-response analysis~\cite{wu2022integrating,wu2025mri,yankeelov2013clinically,chaudhuri2023predictive}. Our study uses patient-derived anatomy and initialization with synthetic longitudinal trajectories to evaluate forecasting, adaptation, and treatment scheduling under controlled conditions.}

\subsection{AI-Augmented Mechanistic Models}

AI-augmented mechanistic models combine explicit physical or biological structure with the flexibility of data-driven learning. Rather than learning the full dynamics from data alone, these approaches retain a mechanistic model for known system behavior and use a learned component to represent unresolved processes, model-form error, or other discrepancies not captured by the governing equations~\cite{raissi2019physics,karniadakis2021physics,willard2022integrating}. Related scientific machine-learning approaches, including sparse model discovery, neural differential equations, and universal differential equations, similarly extend dynamical models with data-driven components while preserving mechanistic structure~\cite{brunton2016discovering,rudy2017data,chen2018neural,rackauckas2020universal}. 
% This strategy is particularly relevant to tumor DTs, where mechanistic models provide interpretable tumor-evolution dynamics but may not fully represent spatial heterogeneity or treatment-conditioned behavior. Previous studies have combined mechanistic tumor-growth models with machine learning for personalized prediction under limited observations~\cite{mascheroni2021improving} and have used deep learning to estimate patient-specific RD parameters from medical imaging~\cite{martens2022deep}. In the present work, the learned component plays a different role. Instead of replacing the mechanistic model or estimating its parameters directly, a residual network learns the structured prediction error left by the RD backbone. The residual therefore represents model-form correction and is not interpreted as an independently identified biological mechanism.
Previous tumor studies combined mechanistic models with machine learning for personalized prediction~\cite{mascheroni2021improving} or used deep learning to estimate RD parameters from imaging~\cite{martens2022deep}. Here, the network learns an additive correction to RD predictions rather than estimating RD parameters. The correction represents model discrepancy, not an independently identified biological mechanism.

\subsection{Recursive Prediction and Adaptive DT Updating}

Accurate one-step prediction does not necessarily lead to stable long-horizon forecasting. During recursive deployment, each predicted state is used as the input for the next prediction, exposing the model to a state distribution that differs from the reference-conditioned data used during training. This discrepancy, commonly associated with exposure bias and rollout distribution shift, can cause local errors to accumulate over time~\cite{bengio2015scheduled}. Adaptive updating provides a way to keep the DT aligned with newly observed patient states. State-estimation methods such as Kalman filtering and ensemble Kalman filtering similarly use incoming observations to correct model forecasts~\cite{kalman1960new,evensen2004sampling}. In contrast, continual and test-time adaptation methods update predictive models as the data distribution evolves~\cite{wang2022continual}. In the proposed framework, newly available observations are used to update the residual network during recursive forecasting while the RD backbone remains fixed. 
% This preserves the mechanistic structure of the model while allowing the learned correction to adapt to the evolving patient-specific trajectory.

\subsection{Treatment Scheduling with Predictive DTs}

Treatment scheduling requires a predictive model that can evaluate how tumor evolution may change under candidate future actions. Dynamical tumor models have long been used in optimal-control and adaptive-therapy studies to investigate treatment timing, dose selection, and response-guided therapy~\cite{martin1992optimal,gatenby2009adaptive,zhang2017integrating}. MPC is well suited to this setting because it repeatedly evaluates candidate treatment sequences over a finite horizon, applies the first selected action, updates the system state, and replans as the trajectory evolves~\cite{mayne2000survey,rawlings2017model,hirata2014model}. Within a tumor DT, MPC therefore provides a direct connection between prediction and decision making. Its performance depends on the accuracy and stability of the predictive model used within the controller, because errors in tumor burden, morphology, or treatment response can alter the ranking of candidate schedules. In this study, the adaptively updated DT is coupled with scenario-based MPC to evaluate chemotherapy and radiotherapy action policies subject to timing, dose, toxicity, and uncertainty constraints. 
% The resulting schedules are used to study the behavior of the integrated prediction–decision framework within the controlled validation setting rather than as clinical treatment recommendations.
% \add{Optimal-control and adaptive-therapy studies have examined treatment timing, dose, and response-guided intervention~\cite{martin1992optimal,gatenby2009adaptive,zhang2017integrating}. MPC repeatedly selects a treatment sequence, applies its first action, and replans from an updated state~\cite{mayne2000survey,rawlings2017model,hirata2014model}. Here, the adapted RD--residual model supplies the forecasts for scenario-based MPC under treatment and toxicity constraints.}

\section{Methods}
The proposed AI-augmented DT framework consists of three main components: (1) an AI-augmented RD model for tumor forecasting, (2) patient-specific adaptation during recursive forecasting, and (3) MPC-based treatment scheduling. The specific model configuration, training procedure, and experimental validation setting are described in Section~\ref{sec:experimental_setting}.

\subsection{Hybrid RD--Residual Learning Model for Tumor Forecasting}
\label{subsec:reference_baseline_rd}

Let $\Omega \subset \mathbb{R}^3$ denote the 3D computational brain domain. 
For each patient, imaging-derived anatomical masks, tissue maps, and tumor segmentation are used to construct the computational domain and initialize the tumor state. 
The brain mask $M_{\mathrm{brain}}$ defines the valid domain, while white- and gray-matter maps determine tissue-dependent diffusion. 
The initial tumor support is denoted by $M_{\mathrm{tumor}}$. 
Tumor state is represented by a normalized cellularity-like field,
\begin{equation}
N(\mathbf{x},t)\in[0,1], 
\qquad 
\mathbf{x}\in\Omega,
\label{eq:tumor_state}
\end{equation}
where $N(\mathbf{x},t)$ denotes normalized tumor cellularity. Total tumor burden is $B(t)=\sum_{\mathbf{x}\in\Omega}N(\mathbf{x},t)\Delta V$, where $\Delta V$ is the voxel volume.

The tumor state at $t=0$ is initialized from a cellularity proxy derived from apparent diffusion coefficient (ADC) maps 
\begin{equation}
N_{\mathrm{ADC}}(\mathbf{x})
=
\left|
\frac{ADC_{\mathrm{ref}}-ADC(\mathbf{x})}
{ADC_{\mathrm{ref}}}
\right|,
\label{eq:adc_proxy}
\end{equation}
where $ADC_{\mathrm{ref}}$ is the ADC normalization value, whose construction is specified in Section 4.1.  
The percentile-based estimate reduces sensitivity to isolated ADC outliers compared with using the maximum value. 
The ADC mapping is used only for initialization. 
The resulting initial tumor state is
\begin{equation}
N(\mathbf{x},0)
=
M_{\mathrm{brain}}(\mathbf{x})
M_{\mathrm{tumor}}(\mathbf{x})
\Pi_{[0,1]}
\left(
N_{\mathrm{ADC}}(\mathbf{x})
\right),
\label{eq:initial_state}
\end{equation}
where $\Pi_{[0,1]}(\cdot)$ denotes clipping to $[0,1]$. 
The tumor state is set to zero outside the reconstructed brain domain.

We model tumor evolution by using an RD equation with tissue-dependent diffusion, logistic proliferation, and treatment-induced tumor reduction:
\begin{equation}
\frac{\partial N(\mathbf{x},t)}{\partial t}
=
\nabla\cdot
\left(
D(\mathbf{x})\nabla N(\mathbf{x},t)
\right)
+
kN(\mathbf{x},t)
\left(
1-\frac{N(\mathbf{x},t)}{\theta}
\right)
-
\Gamma_C(\mathbf{x},t)
-
\Gamma_R(\mathbf{x},t),
\label{eq:rd_baseline}
\end{equation}
where $D(\cdot)$ is the tissue-dependent diffusion field, $k$ is the proliferation parameter, $\theta$ is the carrying-capacity parameter, and $\Gamma_C$ and $\Gamma_R$ denote the effective chemotherapy and radiotherapy responses, respectively. The diffusion field is defined as
\begin{equation}
D(\mathbf{x})
=
D_w M_w(\mathbf{x})
+
D_g M_g(\mathbf{x}),
\label{eq:diffusion_field}
\end{equation}
where $M_w$ and $M_g$ denote the white- and gray-matter maps and $D_w$ and $D_g$ are the corresponding diffusion coefficients.

Let $a_j=(a_j^c,a_j^r)$ denote the treatment action applied at time step $j$, where $a_j^c$ and $a_j^r$ represent the chemotherapy and radiotherapy components, respectively.
During tumor forecasting and patient-specific adaptation, $a_j$ is treated as a known conditioning input. We represent treatment effects through decaying exposure states. 
The post-action chemotherapy and radiotherapy exposure states are
\begin{equation}
c_j^{+}
=
c_j e^{-\lambda_C\Delta t}
+
\eta_C a_j^c,
\qquad
q_j^{+}
=
q_j e^{-\lambda_R\Delta t}
+
a_j^r,
\label{eq:treatment_exposure_states}
\end{equation}
with corresponding tumor-reduction terms
\begin{equation}
\Gamma_C(\mathbf{x},t_j)
=
\beta_C c_j^{+}N(\mathbf{x},t_j),
\qquad
\Gamma_R(\mathbf{x},t_j)
=
\beta_R q_j^{+}N(\mathbf{x},t_j).
\label{eq:treatment_kill_terms}
\end{equation}
The post-action exposure states are carried forward to the next time step, thereby retaining the effects of prior treatment.

Let $f_{\mathrm{RD}}$ denote the time-discretized transition operator associated with Eq.~\eqref{eq:rd_baseline}.  
For compactness, the exposure states and their deterministic updates are included in the transition operator notation. 
The one-step RD prediction is therefore written as
\begin{equation}
N_{j+1}^{\mathrm{RD}}
=
f_{\mathrm{RD}}(N_j,a_j).
\label{eq:rd_transition}
\end{equation}
% \delete{The RD model serves as the mechanistic backbone of the DT.}
% \delete{Its parameters and anatomical inputs describe the prescribed tumor-evolution dynamics, while discrepancies not captured by this model are addressed through residual learning.}
Let $Y_{j+1}$ denote the available target tumor state associated with the transition from $N_j$ under treatment action $a_j$. 
The residual target is defined as
\begin{equation}
r_j^{\mathrm{res}}
=
Y_{j+1}
-
N_{j+1}^{\mathrm{RD}},
\label{eq:residual_target}
\end{equation}
where 
$N_{j+1}^{\mathrm{RD}}
=
f_{\mathrm{RD}}(N_j,a_j)$.

A 3D residual network $f_{\phi}$ is trained to approximate this discrepancy:
\begin{equation}
f_{\phi}(N_j,a_j)
\approx
r_j^{\mathrm{res}}.
\label{eq:residual_model}
\end{equation}
The hybrid RD--residual prediction is then
\begin{equation}
\widehat N_{j+1}
=
\Pi_{[0,1]}
\left[
M_{\mathrm{brain}}
\left(
N_{j+1}^{\mathrm{RD}}
+
f_{\phi}(N_j,a_j)
\right)
\right].
\label{eq:hybrid_prediction}
\end{equation}
The RD component provides the explicit tumor-evolution dynamics, while the residual network corrects structured prediction errors not captured by the mechanistic model.
The residual term is therefore interpreted as model-form correction.

The residual network is first trained offline on one-step transitions using $Y_{j+1}$ as the target tumor state.  
To focus learning on anatomically valid regions and emphasize locations where the RD model exhibits non-negligible prediction error, we define the voxel-wise weight
\begin{equation}
W_j(\mathbf{x})
=
M_{\mathrm{brain}}(\mathbf{x})
\left[
1+
\alpha
I
\left(
\left|r_j^{\mathrm{res}}(\mathbf{x})\right|
>
\tau_{\mathrm{res}}
\right)
\right],
\label{eq:residual_weight}
\end{equation}
where $\alpha$ controls the additional weight assigned to voxels with appreciable residual error, $\tau_{\mathrm{res}}$ is the corresponding residual-magnitude threshold, and $I(\cdot)$ denotes the indicator function. For a training set
$\mathcal D=\{(N_j,a_j,Y_{j+1})\}$,
the offline residual-learning objective is
\begin{equation}
\mathcal L_{\mathrm{offline}}(\phi)
=
\frac{1}{|\mathcal D|}
\sum_{(N_j,a_j,Y_{j+1})\in\mathcal D}
\frac{
\displaystyle
\sum_{\mathbf{x}\in\Omega}
W_j(\mathbf{x})
\left[
f_{\phi}(N_j,a_j)(\mathbf{x})
-
r_j^{\mathrm{res}}(\mathbf{x})
\right]^2
}{
\displaystyle
\sum_{\mathbf{x}\in\Omega}
W_j(\mathbf{x})
}.
\label{eq:offline_loss}
\end{equation}
This objective trains the residual network to correct one-step RD prediction errors at the population level.
The network architecture, training configuration, and values of $\alpha$ and $\tau_{\mathrm{res}}$ used in the experiments are specified in Section~4.

\subsection{Patient-Specific Adaptation During Recursive Forecasting}

During recursive forecasting, the hybrid model repeatedly uses its own predicted tumor state as the input for the next step:
\begin{equation}
\widehat N_{j+1}
=
\Pi_{[0,1]}
\left[
M_{\mathrm{brain}}
\left(
f_{\mathrm{RD}}(\widehat N_j,a_j)
+
f_{\phi}(\widehat N_j,a_j)
\right)
\right].
\label{eq:recursive_prediction}
\end{equation}
Because prediction errors are propagated through subsequent transitions, the forecast may gradually drift away from the evolving patient state.
To reduce this drift, the residual network is adapted when new patient-specific observations become available, while the RD backbone remains fixed.

For each patient, an adaptation buffer stores recent predicted-state transitions together with the applied treatment action and the corresponding observed tumor state:
\begin{equation}
\mathcal B
=
\left\{
\left(
\widehat N_j,
a_j,
Y_{j+1}
\right)
\right\}_{j\in\mathcal I},
\label{eq:adaptation_buffer}
\end{equation}
where $Y_{j+1}$ denotes an available tumor-state observation at time step $j+1$.
Using the recursively predicted state $\widehat N_j$ as the transition input exposes the adaptation procedure to the state distribution encountered during deployment, rather than only to the one-step transitions used during offline training.

Each adaptation update uses $H$ consecutive model time steps ending at the latest observation. Only observed targets contribute to the loss. Starting from the predicted state at the beginning of the window, the hybrid model is propagated using the recorded treatment actions:
\begin{equation}
\widehat N_{j+i+1}
=
\Pi_{[0,1]}
\left[
M_{\mathrm{brain}}
\left(
f_{\mathrm{RD}}
(\widehat N_{j+i},a_{j+i})
+
f_{\phi}
(\widehat N_{j+i},a_{j+i})
\right)
\right],
\qquad
i=0,\ldots,H-1.
\label{eq:recursive_rollout}
\end{equation}
Observations need not be available at every time step.
Let 
$\mathcal O_j
\subseteq
\{1,\ldots,H\}$ 
denote the offsets within the rollout window at which tumor-state observations are available. 
The residual-network parameters are updated by minimizing the masked rollout loss
\begin{equation}
\mathcal L_{\mathrm{adapt}}(\phi)
=
\frac{1}{|\mathcal O_j|}
\sum_{i\in\mathcal O_j}
\mathcal L_{\mathrm{mask}}
\left(
\widehat N_{j+i},
Y_{j+i}
\right),
\label{eq:online_loss}
\end{equation}
where $\mathcal L_{\mathrm{mask}}$ denotes a voxel-wise reconstruction loss over the valid brain domain. 
Unlike one-step offline training, this objective directly penalizes errors accumulated during recursive forecasting.
% \delete{Updating only the residual component allows the DT to adapt to newly observed patient-specific behavior while preserving the mechanistic RD dynamics.}

\subsection{MPC-Based Treatment Scheduling}
\label{subsec:mpc}

During tumor forecasting and patient-specific adaptation, the treatment action
$a_j$ is treated as a known conditioning input. 
For treatment scheduling, future treatment actions are instead treated as decision variables. 
Let $u_j=(u_j^c,u_j^r)$ denote a candidate chemotherapy and radiotherapy action at decision step $j$, and let
$H_{\mathrm{MPC}}$ denote the planning horizon. The adapted DT serves as the forward model for evaluating candidate treatment sequences. 
For a candidate action $u_j$, the predicted tumor state evolves according to
\begin{equation}
N_{j+1}
=
\Pi_{[0,1]}
\left[
M_{\mathrm{brain}}
\left(
f_{\mathrm{RD}}(N_j,u_j)
+
f_{\phi}(N_j,u_j)
\right)
\right].
\label{eq:mpc_dynamics}
\end{equation}
 During MPC rollout, residual corrections with magnitude below $\epsilon_r$ are set to zero before masking and clipping.

Treatment effects during MPC rollout follow the exposure-state dynamics introduced in Section~\ref{subsec:reference_baseline_rd}. 
For a candidate action $u_j$, the post-action exposure states are
\begin{equation}
c_j^{+,\mathrm{MPC}}
=
c_j^{\mathrm{MPC}}
e^{-\lambda_C\Delta t}
+
\eta_C u_j^c,
\qquad
q_j^{+,\mathrm{MPC}}
=
q_j^{\mathrm{MPC}}
e^{-\lambda_R\Delta t}
+
u_j^r,
\label{eq:mpc_exposure_states}
\end{equation}
with corresponding treatment-induced tumor-reduction terms
\begin{equation}
\Gamma_C^{\mathrm{MPC}}(\mathbf{x},t_j)
=
\beta_C
c_j^{+,\mathrm{MPC}}
N_j(\mathbf{x}),
\qquad
\Gamma_R^{\mathrm{MPC}}(\mathbf{x},t_j)
=
\beta_R
q_j^{+,\mathrm{MPC}}
N_j(\mathbf{x}).
\label{eq:mpc_treatment_terms}
\end{equation}
The resulting exposure states are propagated throughout each candidate rollout so that prior treatment influences subsequent predicted tumor evolution.

Treatment modality, timing, dose, and toxicity constraints define the feasible action set
$\mathcal U_j$ at each decision step. Denote a candidate treatment sequence over the planning horizon by 
$
\mathbf{u}_{j:j+H_{\mathrm{MPC}}-1}
=
(u_j,u_{j+1},\ldots,u_{j+H_{\mathrm{MPC}}-1}).
$ 
Because exhaustive optimization over all possible treatment sequences is computationally expensive, the controller uses random shooting to sample a finite set of feasible candidate sequences,
\begin{equation}
\mathcal U_j^{\mathrm{shoot}}
\subset
\mathcal U_j
\times
\mathcal U_{j+1}
\times
\cdots
\times
\mathcal U_{j+H_{\mathrm{MPC}}-1}.
\label{eq:sampled_action_set}
\end{equation}
The selected sequence is therefore the best schedule identified among the sampled candidates rather than a guaranteed global optimum over the full action space.

\begin{algorithm}
\caption{AI-augmented DT prediction--scheduling loop}
\label{alg:ai_augmented_dt_loop}
\begin{algorithmic}[1]
\Require Initial state $N_0$, RD transition $f_{\mathrm{RD}}$, residual model $f_{\phi}$, feasible action sets $\{\mathcal{U}_j\}$, observations $\{Y_j\}$, planning horizon $H_{\mathrm{MPC}}$, adaptation rollout window $H$, minimum buffer size $B_{\min}$
\State Initialize $\widehat{N}_0\leftarrow N_0$, adaptation buffer $\mathcal{B}\leftarrow\emptyset$, exposure states $c_0^{\mathrm{MPC}}\leftarrow0$, $q_0^{\mathrm{MPC}}\leftarrow0$, and toxicity state $\mathrm{tox}_0\leftarrow0$
\For{$j=0,\ldots,T-1$}
    \If{a new observation $Y_j$ is available and $j>0$}
        \State Add the recent transition $(\widehat{N}_{j-1},a_{j-1},Y_j)$ to $\mathcal{B}$
        \If{$|\mathcal{B}|\geq B_{\min}$}
            \State Update the residual parameters $\phi$ using Eq.~\eqref{eq:online_loss}
        \EndIf
    \EndIf
    \State Sample feasible candidate schedules $\mathcal{U}^{\mathrm{shoot}}_j$ using Eq.~\eqref{eq:sampled_action_set}
    \State Roll out each candidate using Eq.~\eqref{eq:mpc_dynamics} under the perturbation scenarios in Eq.~\eqref{eq:mpc_initial_scenarios}
    \State Select $\mathbf{u}^{*}_{j:j+H_{\mathrm{MPC}}-1}$ by minimizing Eq.~\eqref{eq:mpc_objective}
    \State Execute the first action $u_j^{*}$ and set $a_j\leftarrow u_j^{*}$
    \State Update exposure states using Eq.~\eqref{eq:mpc_exposure_states} and toxicity using Eq.~\eqref{eq:toxicity} with $u_j=a_j$
    \State Forecast $\widehat{N}_{j+1}$ using Eq.~\eqref{eq:mpc_dynamics} with $u_j=a_j$
\EndFor
\end{algorithmic}
\end{algorithm}

To account for uncertainty in the current tumor-state estimate, each candidate sequence is evaluated under bounded perturbations of the current DT state,
\begin{equation}
N_j^{(\sigma)}
=
\Pi_{[0,1]}
\left[
M_{\mathrm{brain}}
(1+\sigma e_{\max})
\widehat N_j
\right],
\qquad
\sigma\in\{-1,+1\},
\label{eq:mpc_initial_scenarios}
\end{equation}
where $e_{\max}$ controls the perturbation magnitude. 
For each candidate sequence, the DT is propagated over the planning horizon under both perturbation scenarios. 
The controller selects the sequence that minimizes the worst-case terminal tumor burden,
\begin{equation}
\mathbf{u}_{j:j+H_{\mathrm{MPC}}-1}^{*}
=
\arg\min_{\mathbf{u}\in\mathcal U_j^{\mathrm{shoot}}}
\;
\max_{\sigma\in\{-1,+1\}}
\left[
\sum_{\mathbf{x}\in\Omega}
N_{j+H_{\mathrm{MPC}}}^{(\sigma)}
(\mathbf{x};\mathbf{u})
\Delta V
\right],
\label{eq:mpc_objective}
\end{equation}
where $\Delta V$ denotes voxel volume. 
This scenario-based objective accounts for the prescribed perturbations in the current tumor-state estimate but does not provide a formal robustness guarantee against unmodeled sources of uncertainty.

Chemotherapy toxicity is represented by a cumulative state,
\begin{equation}
\mathrm{tox}_{j+1}
=
\mathrm{tox}_{j}e^{-\lambda_{\mathrm{tox}}\Delta t}
+
\beta_{\mathrm{tox}}u^c_j,
\qquad \text{subject to} \qquad
\mathrm{tox}_{j+1}
\leq
\mathrm{tox}_{\max},
\label{eq:toxicity}
\end{equation}
where $\lambda_{\mathrm{tox}}$ controls toxicity decay, $\beta_{\mathrm{tox}}$ controls dose-dependent accumulation, and $\mathrm{tox}_{\max}$ is the allowable toxicity limit. 
Candidate chemotherapy actions that violate this constraint are replaced by no treatment. At each decision step, the controller evaluates the sampled treatment sequences, selects
$\mathbf{u}_{j:j+H_{\mathrm{MPC}}-1}^{*}$, and executes only the first action $u_j^*$. 
Once executed, the selected control becomes the applied treatment action, $a_j = u_j^*$. 
The DT state and treatment-exposure states are then advanced using this applied action, and the optimization is repeated at the next decision step. 
This receding-horizon procedure allows the treatment schedule to change as the predicted tumor state evolves and new patient-specific observations become available. 

Algorithm~\ref{alg:ai_augmented_dt_loop} summarizes the closed-loop prediction, updating, and treatment-scheduling procedure.

% The MPC-generated schedules are interpreted as computational treatment-action policies within the controlled evaluation setting, not as clinically actionable treatment recommendations.

\section{Experimental Setting}
\label{sec:experimental_setting}
\subsection{Patient-Informed Synthetic Validation Dataset}

Experiments were conducted using synthetic longitudinal tumor trajectories initialized from patient-specific clinical imaging.
Anatomical brain masks, tissue maps, ADC images, and tumor regions were derived from the publicly available UPENN-GBM cohort \cite{bakas2022university}.
Because frequently sampled longitudinal imaging under controlled treatment conditions is rarely available at cohort scale, clinical imaging was used to define patient-specific anatomy and tumor initialization, while subsequent tumor evolution was generated computationally.
Throughout this section and the Results, ``reference'' denotes a simulator-generated tumor trajectory.

For each patient, the initial tumor state was constructed using the formulation in Section~3.1.
In the present implementation, $ADC_{\mathrm{ref}}$ was estimated as the 99th percentile of valid ADC values within the tumor region, with the brain mask used as a fallback when the tumor region was too small.
All patient-specific fields were aligned to a common voxel space and resampled to a $128\times128\times128$ computational grid. Discrete masks were resampled using nearest-neighbor interpolation, whereas continuous imaging-derived quantities were resampled using trilinear interpolation.

% For each patient, the controlled reference simulator generated a 120-step tumor-evolution trajectory. In the current parameterization, one simulation step corresponds to one model-integration day; however, the experiments are reported primarily in simulation steps to avoid implying that clinical observations are available at this frequency. This produced 120 one-step transitions and 121 stored tumor states, including the initialized state at step~0. All patient-specific fields were aligned to a common voxel space and resampled to a unified computational grid before simulation. Discrete masks were resampled using nearest-neighbor interpolation, whereas continuous imaging-derived quantities were resampled using trilinear interpolation. A total of 387 synthetic tumor trajectories were generated. Data were partitioned at the patient level into training, validation, and testing cohorts to prevent leakage of patient-specific anatomy or tumor initialization between model development and final evaluation. The dataset and prediction settings are summarized in Table~\ref{tab:exp_setup}. 

Each patient-specific initial tumor state was propagated for 120 simulation steps, producing 120 one-step transitions and 121 stored states, including the initial state at step 0. Under the present parameterization, one simulation step corresponds to one model-integration day. The term ``simulation step'' is used throughout the experimental analysis to distinguish this numerical resolution from the frequency of clinically available observations. A total of 387 patient-informed synthetic trajectories were generated.
The data were partitioned at the patient level into training, validation, and testing cohorts using an 80\%/10\%/10\% split.
Patient-level separation ensured that anatomy and tumor initialization from a given patient appeared in only one partition.

Each trajectory contained both tumor states and the treatment actions applied during trajectory generation.
Treatment histories used for residual-model development were generated in separate controller runs and were not derived from the held-out treatment-scheduling experiments.
The resulting transitions had the form 
$(N_j,a_j,N_{j+1})$, 
where $a_j$ denotes the treatment action applied at simulation step $j$.
The detailed action space and feasibility constraints used to generate treatment-conditioned trajectories are described in Section~4.5. 
The principal dataset and prediction settings are summarized in Table~\ref{tab:exp_setup}.

\begin{table}
\centering
\caption{Summary of dataset and prediction settings.}
\label{tab:exp_setup}
\begin{tabularx}{\linewidth}{
    >{\raggedright\arraybackslash}p{0.32\linewidth}
    >{\raggedright\arraybackslash}X
}
\toprule
Item & Value \\
\midrule
Patients / trajectories & 387 \\
Trajectory length & 120 simulation steps; 121 stored states including step~0 \\
Temporal resolution & 1 model-integration day per simulation step \\
Spatial grid & $128 \times 128 \times 128$ \\
Train / validation / test split & 80\% / 10\% / 10\% at the patient level \\
Forecast rollout horizon & 120 simulation steps \\
MPC planning horizon & 10 simulation steps \\
Residual-model inputs & Current tumor state and treatment action \\
Compared forecasting models & RD baseline, offline hybrid RD--residual, adapted hybrid RD--residual \\
Hardware & HPC workstation with one NVIDIA H200 GPU \\
\bottomrule
\end{tabularx}
\end{table}

\subsection{Reference Simulator and RD Model Configuration}

A controlled reference simulator was used to generate longitudinal tumor states for model development and evaluation.
The reference simulator followed the same general RD and treatment-response structure introduced in Section~3.1, but used a spatially varying proliferation field:
\begin{equation}
\frac{\partial N(\mathbf{x},t)}{\partial t}
=
\nabla\cdot
\left(
D(\mathbf{x})\nabla N(\mathbf{x},t)
\right)
+
k(\mathbf{x})
N(\mathbf{x},t)
\left(
1-\frac{N(\mathbf{x},t)}{\theta}
\right)
-
\Gamma_C(\mathbf{x},t)
-
\Gamma_R(\mathbf{x},t).
\end{equation}
The RD model used within the proposed DT retained the formulation in Section~3.1 with a spatially constant proliferation parameter $k$.
This difference introduced a controlled source of model-form discrepancy for evaluating the learned residual correction.

Both models were advanced using a unit simulation time step, $\Delta t=1$, with carrying capacity normalized to $\theta=1$. 
The tissue-dependent diffusion field used $D_g=0.03$ in gray matter and $D_w=5D_g$ in white matter.
The DT RD model used $k=0.15$, whereas the reference simulator used a spatially varying proliferation field with a white-matter-to-base proliferation ratio of 1.5.

Treatment effects in the reference simulator and DT model followed the same exposure-state dynamics.
The chemotherapy parameters were
$\beta_C=0.03$, $\lambda_C=0.30$, and $\eta_C=1$, and the radiotherapy parameters were
$\beta_R=0.30$ and $\lambda_R=0.50$.
Using the same treatment-response formulation in both models ensured that the controlled discrepancy arose primarily from the proliferation representation rather than from differences in treatment-effect equations.

For residual-model training, the target state $Y_{j+1}$ introduced in Section~3.1 was instantiated as the corresponding simulator-generated reference state, 
$Y_{j+1}=N_{j+1}^{\mathrm{ref}}$. 
The residual-learning target was therefore the discrepancy between the controlled reference transition and the RD prediction under the same applied treatment action.

\subsection{Residual Network and Offline Training}

The residual correction model was implemented as a 3D U-Net based on established volumetric U-Net architectures \cite{ronneberger2015u,cciccek20163d,isensee2021nnu}.
The network received three input channels: the current tumor state, a chemotherapy-action map, and a radiotherapy-action map.
The two treatment-action maps were spatially constant over the computational domain and encoded the applied action $a_j$.
Chemotherapy and radiotherapy values were normalized by their respective maximum values in the training set before being expanded into spatially constant maps.
The network produced a voxel-wise residual field with the same spatial dimensions as the tumor state.

The encoder used feature widths of 24, 48, and 96, followed by a 192-channel bottleneck.
The decoder used trilinear upsampling and skip connections.
Residual convolutional blocks used instance normalization and LeakyReLU activation.
A dropout rate of 0.1 was applied at the bottleneck, and the network output was multiplied by a learnable residual scale initialized to 0.1. 
Offline residual learning used the weighted objective defined in Section~3.1.
The residual-weighting parameters were set to
$\alpha=10$ and $\tau_{\mathrm{res}}=10^{-4}$.
Training used $32\times32\times32$ 3D patches sampled from treatment-conditioned trajectories.
The model was optimized using AdamW for 200 epochs with a batch size of 12, learning rate $10^{-4}$, and weight decay $10^{-5}$.
The gradient norm was clipped at 1.0.

\subsection{Forecasting and Adaptation Protocol}

Forecasting performance was evaluated using three model configurations.
The RD baseline used only the mechanistic transition $f_{\mathrm{RD}}$.
The offline hybrid model combined the RD prediction with the learned residual correction but remained fixed during recursive forecasting.
The online-adapted hybrid model, hereafter referred to as the adapted hybrid model, updated the residual-network parameters using newly available patient-specific observations while keeping the RD backbone fixed. 
In all forecasting experiments, treatment actions were known inputs taken from the corresponding synthetic trajectory rather than decision variables.

Offline residual-correction accuracy was first evaluated using held-out reference-conditioned transitions.
For these experiments, dense and sparse sampling refer to the stride between the starting time points of sampled prediction pairs. Every pair consists of a reference input state $Y_j$ and its immediately subsequent target state $Y_{j+1}$, so the prediction horizon remains one simulation step in both settings. These sampling strides are distinct from the frequency of observations provided for patient-specific adaptation. 
Recursive forecasting was evaluated separately.
Each model was initialized from the tumor state at step~0 and propagated over the full 120-step horizon, with each predicted state used as the input to the next transition.
Predictions were compared with the corresponding reference states at configuration-specific evaluation time points. The same RD trajectory was used in both configurations; the different RD baseline summaries arose solely from aggregating prediction errors over different evaluation time points.

% \delete{For patient-specific adaptation, the rollout horizon in Eq.~\eqref{eq:online_loss} was set to $H=3$ simulation steps. In the primary dense-observation setting, a reference state was made available at every simulation step.
% Adaptation began after five transitions had accumulated, and the buffer retained at most 60 recent transitions. Following each new observation, the residual network was updated for 10 optimization steps using a learning rate of $10^{-5}$. This setting was designed to characterize adaptation under the most frequent observation schedule considered in the study.} 
Adaptation used an $H=3$-step rollout window and began after $B_{\min}=5$ observed transitions. The buffer held up to 60 observed transitions. In the dense setting, observations were available at every step, followed by 10 optimization steps at a learning rate of $10^{-5}$.

Observation-frequency sensitivity was evaluated separately by providing reference observations every seven simulation steps.
Adaptation was performed only when a scheduled observation became available, and unobserved states were excluded from the adaptation loss.
In both observation settings, the available patient-specific observation $Y_j$ corresponded to the simulator-generated reference state at the same step.

\subsection{Treatment-Scheduling Protocol}

Treatment scheduling was evaluated by coupling the adapted hybrid DT with the scenario-based MPC formulation described in Section~\ref{subsec:mpc}.
The full tumor trajectory extended over 120 simulation steps, whereas MPC optimized candidate treatment sequences over a local planning horizon of
$H_{\mathrm{MPC}}=10$ steps.
The controller therefore repeatedly performed finite-horizon optimization while the overall tumor trajectory continued over the 120-step evaluation period. Treatment steps are numbered from 1 to 120 in the experimental descriptions and figures. Treatment step $s$ corresponds to action $a_{s-1}$ in the transition from $N_{s-1}$ to $N_s$ in Algorithm~\ref{alg:ai_augmented_dt_loop}.

At each decision step, random shooting generated 120 candidate treatment sequences.
Treatment actions were mutually exclusive and selected from
\[
\{\mathrm{none},\mathrm{chemotherapy},\mathrm{radiotherapy}\}.
\]
Treatment was eligible every seven simulation steps beginning at step~1.
At eligible steps, chemotherapy doses were sampled as integer action units from 0 to 25.
Radiotherapy was represented as a 2-Gy action and was permitted at most once over the simulated treatment course.
If a sampled chemotherapy action violated the toxicity constraint, that action was replaced by no treatment.

Candidate MPC trajectories were propagated using the same treatment-exposure dynamics as the DT forecasting model.
During controller rollout, residual corrections with absolute magnitude below
$\epsilon_r=0.1$ were suppressed.
The bounded state-perturbation magnitude in the scenario-based objective was set to
$e_{\max}=0.001$.
For each decision step, the controller selected the sampled sequence that minimized the worst-case terminal tumor burden across the two perturbation scenarios, executed only the first action, and repeated the optimization at the next decision step. 
Chemotherapy toxicity was initialized with
$\mathrm{tox}_0=0$ and used
$\lambda_{\mathrm{tox}}=0.30$,
$\beta_{\mathrm{tox}}=0.40$, and
$\mathrm{tox}_{\max}=5$.

During MPC schedule generation, the reference simulator was advanced under each executed action $a_j$ to produce the next observation $Y_{j+1}$. Observations were supplied at every model time step, and residual-network updates followed the buffer and optimization settings in Section~4.4 before replanning. After schedule generation, the executed action sequence was replayed through the reference simulator to compute the reported treatment outcomes.
For comparison, the fixed synthetic schedule administered one 2-Gy radiotherapy action at step~1, followed by 8 chemotherapy action units every seven simulation steps beginning at step~8. Chemotherapy was therefore administered at steps~8, 15, \ldots, 120, giving 17 administrations and 136 action units in total. No chemotherapy was administered at the initial radiotherapy step.
Both the fixed schedule and MPC schedule satisfied the constraint of administering at most one treatment modality per decision step, and both were evaluated using the same reference dynamics.
This experiment was designed to evaluate treatment-scheduling behavior within the controlled computational setting.

\subsection{Evaluation Metrics}

Forecasting was evaluated within the brain mask using mean squared error (MSE), Dice, peak signal-to-noise ratio (PSNR), and absolute tumor-burden error, $E_{B,j}=|\widehat B_j-B_j^{\mathrm{ref}}|$. 
Recursive forecasting performance was evaluated over the 120-step rollout for the RD baseline, offline hybrid model, and adapted hybrid model.

Treatment-scheduling performance was evaluated after replaying each schedule through the controlled reference simulator.
Reported endpoints included final tumor burden at step~120, cumulative tumor burden over the full trajectory, total chemotherapy exposure, total radiotherapy exposure, maximum toxicity, and the number of toxicity-violation steps.
Final tumor burden measures tumor control at the terminal evaluation point, whereas cumulative burden measures total tumor burden accumulated over the full simulated trajectory.
Both were reported because improvement in terminal burden does not necessarily imply improvement in cumulative burden.

The evaluation metrics used in each experiment are summarized in Table~\ref{tab:metrics}.
Forecasting metrics are reported as mean $\pm$ standard deviation across the held-out test cohort unless otherwise stated.
Treatment-scheduling metrics are reported using cohort-level summary statistics appropriate to the corresponding comparison.

\begin{table}[H]
\centering
\caption{Evaluation metrics used in each experiment.}
\label{tab:metrics}
\begin{tabular}{ll}
\toprule
Objective & Metrics \\
\midrule
Forecasting accuracy & MSE, Dice, PSNR, tumor-burden error \\
Recursive forecasting & MSE, Dice, tumor-burden error over rollout \\
Treatment scheduling & Final tumor burden, cumulative tumor burden, dose exposure, toxicity \\
\bottomrule
\end{tabular}
\end{table}

\section{Results}

\begin{figure}
\centering
\includegraphics[width=\textwidth]{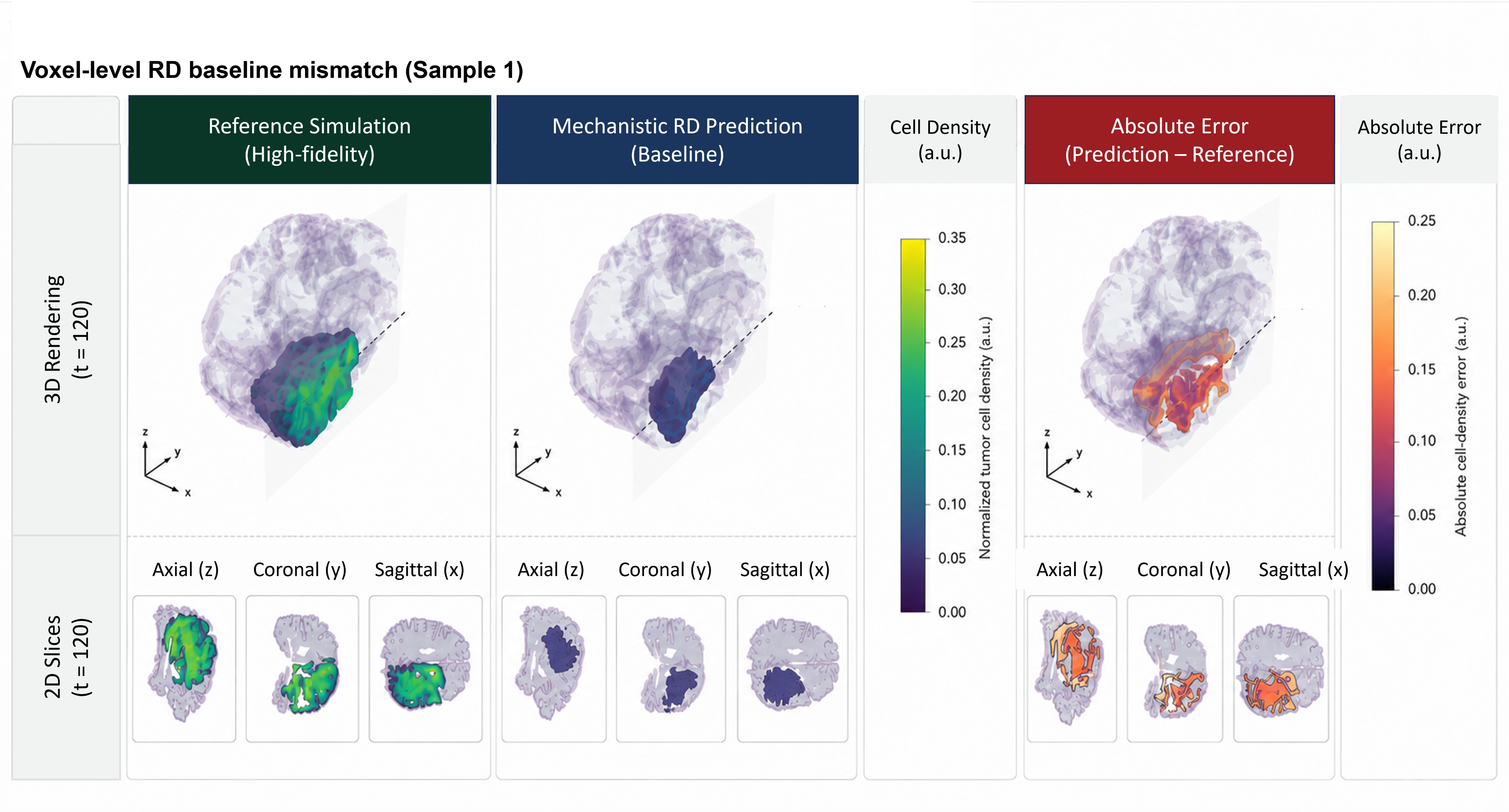}
\caption{
Voxel-level comparison between the controlled reference simulation and baseline RD prediction for an illustrative patient at simulation step~120. The left column shows the reference state, the middle column shows the RD prediction after recursive rollout, and the right column shows absolute voxel-wise prediction error. 3D renderings show global tumor extent, while axial, coronal, and sagittal slices expose internal density mismatch. Shared color scales are used for cellularity and absolute error.
}
\label{fig:baseline_spatial_failure}
\end{figure}

% All experiments were conducted under the controlled synthetic validation setting described in Section \ref{sec:experimental_setting}, using patient-specific imaging-derived anatomy and tumor initialization. 
% Results are organized to match the main components of the proposed framework. We first evaluate the baseline RD model to characterize the controlled mechanistic mismatch created by the model hierarchy. We then assess whether hybrid residual learning corrects this mismatch in one-step forecasting. Next, we evaluate whether patient-specific online updating improves long-horizon recursive forecasting. Finally, we couple the online-adapted model with scenario-based robust MPC to evaluate constrained treatment-scheduling behavior, followed by an observation-frequency sensitivity analysis.
All experiments were conducted using the patient-informed synthetic validation setting described in Section~\ref{sec:experimental_setting}.
We first characterize the forecasting error of the mechanistic RD model, then evaluate residual correction under reference-conditioned prediction, followed by patient-specific adaptation during recursive forecasting.
We subsequently evaluate MPC-based treatment scheduling using the adapted DT and conclude with an analysis of sensitivity to observation frequency.

\subsection{Baseline RD Forecasting Performance}
\label{subsec:mechanistic_baseline}

% The first experiment evaluates the baseline RD model alone. This experiment corresponds to the patient-specific reference and baseline RD models defined in Section~\ref{subsec:reference_baseline_rd}: the reference trajectory is generated by the controlled synthetic simulator, whereas the baseline RD model uses a simplified proliferation representation. The goal is not to compare against observed clinical progression, but to determine whether the baseline mechanistic backbone captures coarse tumor evolution and to identify the structured error that residual learning must correct.

We first evaluated the RD model against the controlled reference trajectories.
The reference simulator incorporated spatially varying proliferation, whereas the RD forecasting model used the spatially simplified formulation described in Section~\ref{sec:experimental_setting}.
This comparison therefore characterizes the structured model-form discrepancy that the learned residual component is designed to correct.

Figure~\ref{fig:baseline_spatial_failure} shows an illustrative patient at the end of the 120-step recursive rollout.
The RD prediction preserves the approximate tumor location and overall spatial extent but underestimates tumor cellularity within the lesion.
The largest errors occur within tumor-bearing regions, including irregular margins and areas of higher cellularity, consistent with a spatially structured model-form discrepancy.

% Figure~\ref{fig:baseline_spatial_failure} shows a representative failure mode of the baseline RD model at the end of the 120-step rollout. The RD prediction preserves the approximate tumor location and coarse spatial footprint, indicating that the mechanistic backbone captures the dominant spatial invasion pattern. However, it substantially underestimates heterogeneous cellularity within the lesion. The error is concentrated within the tumor-bearing region, particularly near irregular margins and high-density areas, indicating structured model-form bias rather than random numerical noise.

\begin{figure}
\centering
\includegraphics[width=1\textwidth]{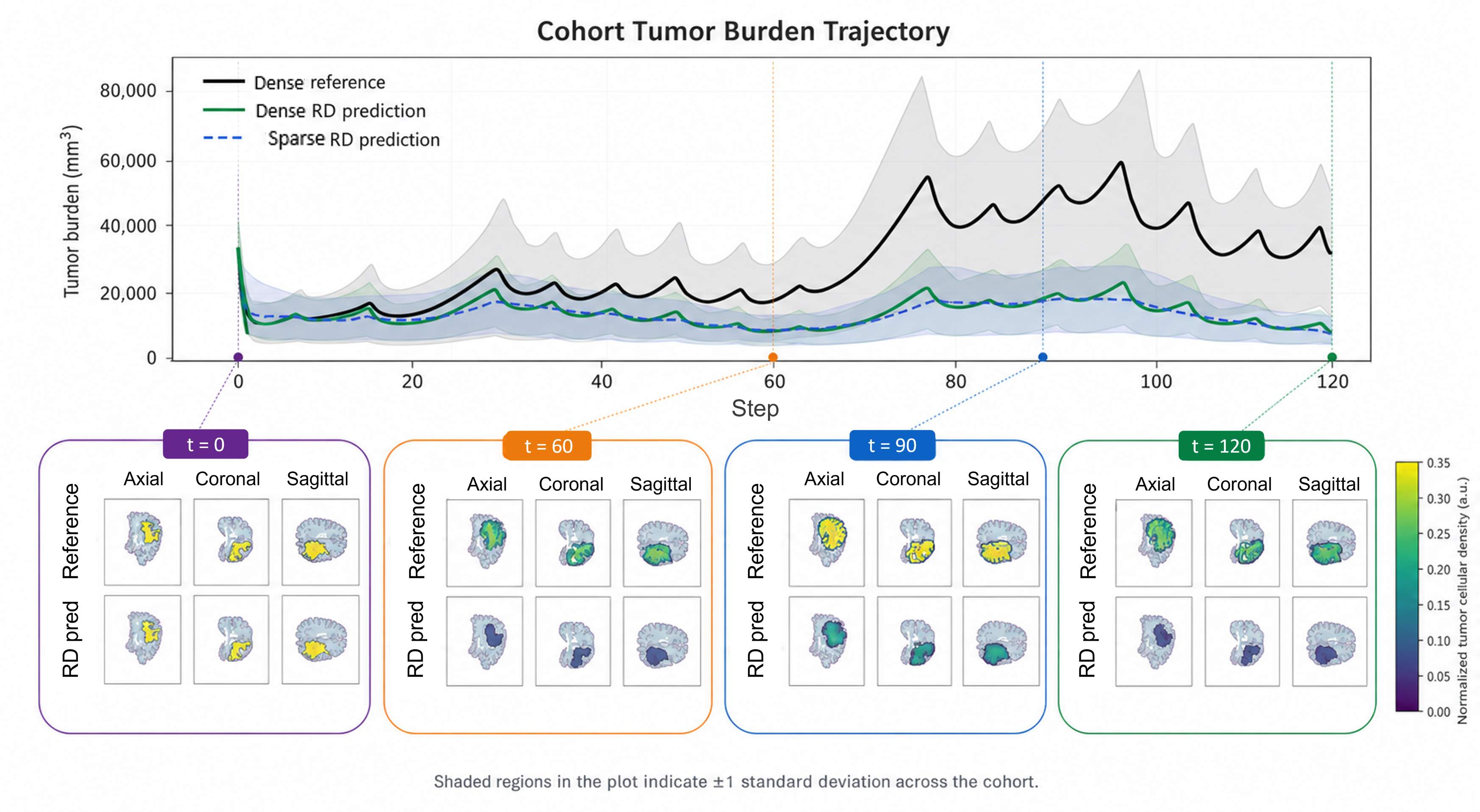}
\caption{
Cohort-level tumor-burden trajectories over the 120-step controlled rollout. The black curve denotes the reference trajectory, while the green and blue curves denote baseline RD predictions evaluated under dense and sparse simulated settings. Shaded regions indicate $\pm 1$ standard deviation across patients. Insets show anatomical slices from an illustrative patient at selected simulation steps, linking temporal burden mismatch to spatial underestimation by the RD model.
}
\label{fig:baseline_tumor_burden}
\end{figure}

% The cohort-level trajectories in Fig.~\ref{fig:baseline_tumor_burden} show that the RD model reproduces the broad temporal behavior of the reference trajectory, including early tumor-burden reduction followed by later regrowth. However, both dense and sparse RD experimental settings increasingly underestimate tumor burden during the later rollout period. The separation becomes pronounced after approximately simulation step~70 and persists through step~120. 
The cohort-level tumor-burden trajectories in Fig.~\ref{fig:baseline_tumor_burden} show that the RD model reproduces the overall temporal pattern of the reference trajectories, including an initial reduction in tumor burden followed by subsequent regrowth.
However, RD predictions increasingly underestimate tumor burden during the later portion of the rollout in both sampling configurations.
The separation from the reference trajectory becomes more pronounced after approximately step~70 and persists through step~120.

\begin{figure}
\centering
\includegraphics[width=\textwidth]{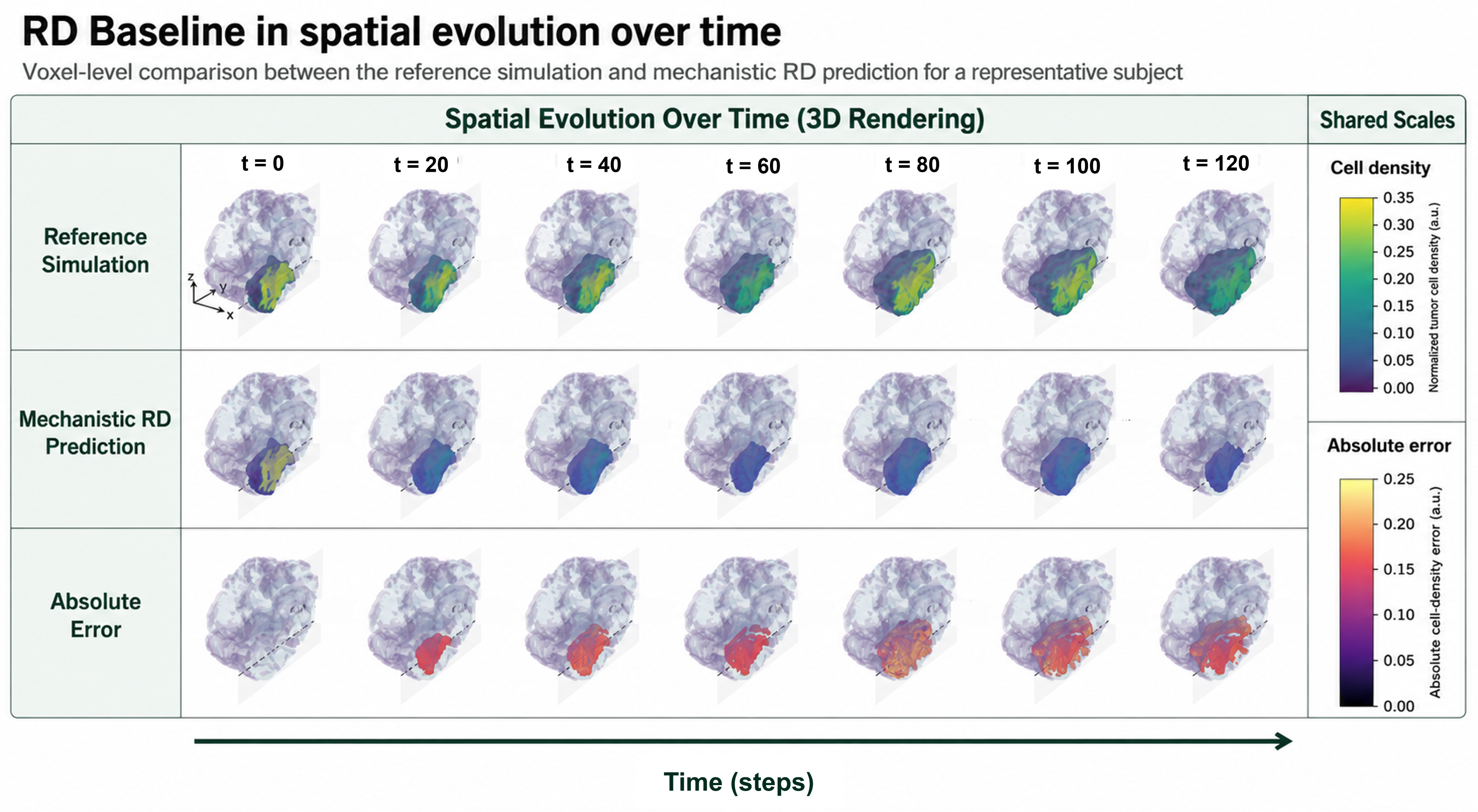}
\caption{
Spatial evolution of baseline RD error over the 120-step rollout for an illustrative patient. Columns show selected time points from simulation step~0 to step~120. Rows show the reference state, baseline RD prediction, and absolute error using shared color scales. Error emerges after initialization and accumulates as the baseline RD model underestimates heterogeneous tumor density during long-horizon recursive prediction.
}
\label{fig:baseline_spatial_evolution_failure}
\end{figure}

% Figure~\ref{fig:baseline_spatial_evolution_failure} further shows that this burden mismatch develops gradually, local density errors appear after initialization and accumulate during recursive forecasting. Figure~\ref{fig:baseline_metrics} quantifies the same trend. Masked voxel-wise MSE increases over time and reaches its largest values during the later simulation period. Tumor-burden error follows a similar pattern, with some patients reaching errors on the order of $10^4$ to $10^5$ mm$^3$. Dice overlap starts near 1.0 at initialization and decreases to approximately 0.35--0.45 by the end of the rollout. The patient-level distributions indicate that this behavior is not limited to a single illustrative case.

\begin{figure}[H]
\centering
\includegraphics[width=\textwidth]{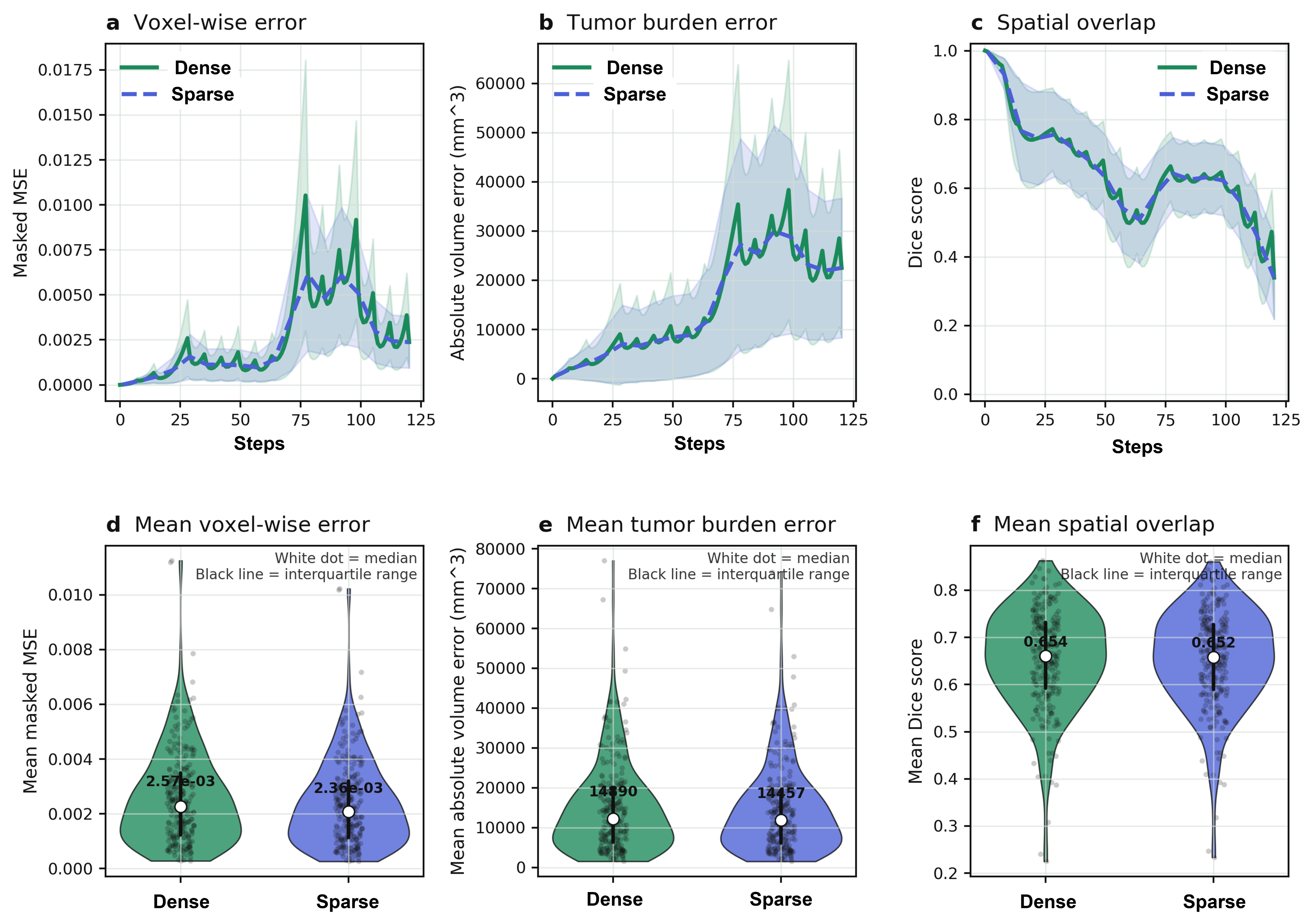}
\caption{
Quantitative evaluation of baseline RD forecasting performance across the cohort. Panels (a)--(c) show temporal evolution of masked voxel-wise MSE, absolute tumor-burden error, and Dice spatial overlap for dense and sparse RD predictions during the 120-step rollout. Panels (d)--(f) summarize patient-level distributions of the corresponding mean metrics. White dots indicate medians, black bars indicate interquartile ranges, and individual points represent patient-level values.
}
\label{fig:baseline_metrics}
\end{figure}

Figure~\ref{fig:baseline_spatial_evolution_failure} illustrates how this discrepancy develops over time for an individual patient.
Prediction error is small near initialization but progressively accumulates during recursive forecasting, particularly within regions of heterogeneous tumor density.
The cohort-level results in Fig.~\ref{fig:baseline_metrics} show the same trend.
Masked voxel-wise MSE and absolute tumor-burden error increase during the rollout, while Dice overlap progressively decreases.
By the end of the rollout, Dice values are approximately 0.35--0.45, and tumor-burden errors reach $10^4$--$10^5$~mm$^3$ for some patients.

These results establish the role of the baseline RD model in the proposed framework. It provides an interpretable mechanistic backbone that captures coarse tumor location and broad temporal direction, but it systematically underestimates heterogeneous tumor density and accumulates burden error during recursive rollout. The next experiment evaluates whether residual learning can correct this structured mismatch while retaining the RD model as the forecasting backbone.

\subsection{Hybrid RD--Residual Forecasting Performance}
\label{subsec:hybrid_forecasting}

% The second experiment evaluates the hybrid RD--residual forecasting model. The purpose is to test whether an additive learned residual can correct the structured voxel-level discrepancy left by the baseline RD model. In this experiment, predictions are compared with the controlled reference trajectory; therefore, the results measure simulator-consistent correction of mechanistic model mismatch rather than clinical biological validation.
We next evaluated whether the learned residual component corrects the structured prediction error of the RD model.
This analysis used held-out reference-conditioned transitions and therefore evaluates one-step model-form correction separately from recursive forecasting.

% Figures~\ref{fig:hybrid_qualitative_sample1} and~\ref{fig:hybrid_qualitative_sample2} show that residual correction substantially improves spatial reconstruction of tumor cellularity in two visually distinct patient samples. The two examples are included to demonstrate that the evaluation cohort contains different tumor shapes and spatial density patterns, rather than a single repeated morphology. In both samples, the RD baseline smooths the tumor field and underestimates high-density regions, consistent with the failure mode identified in Section~\ref{subsec:mechanistic_baseline}. In contrast, the offline hybrid model better preserves tumor extent and internal density variation. The error maps show that large contiguous RD error regions are reduced after residual correction, particularly within the tumor-bearing region.

\begin{figure}[H]
\centering
\includegraphics[width=\textwidth]{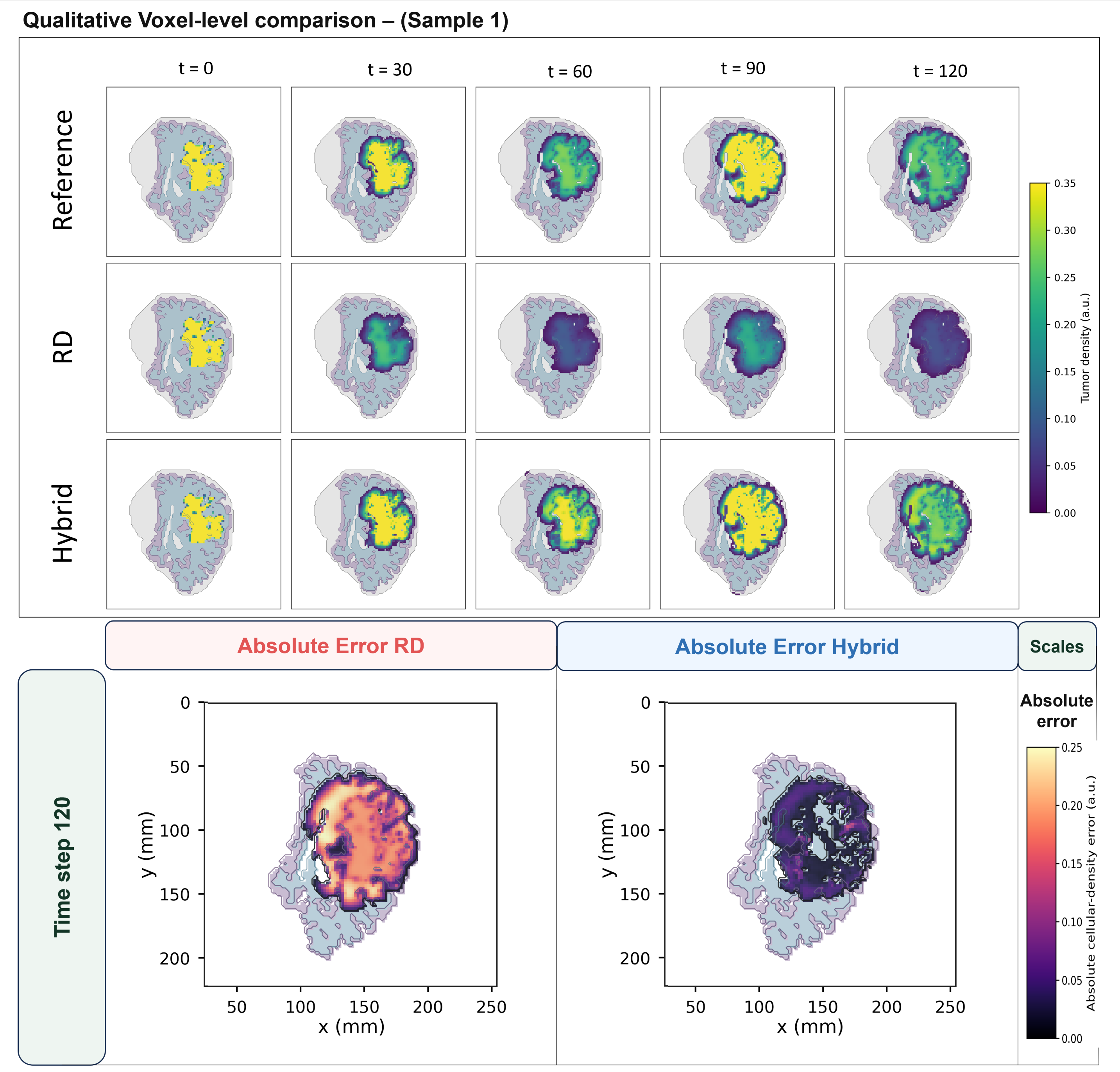}
\caption{
Qualitative voxel-level comparison for patient sample~1 in the dense
forecasting setting. The top panel shows the controlled reference state,
baseline RD prediction, and offline hybrid RD--residual prediction at
selected simulation steps from step~0 to step~120. For each displayed
post-initialization step, the RD and hybrid outputs are one-step predictions conditioned on the corresponding preceding reference state, rather than states generated by a free-running recursive rollout from step~0. The bottom panel compares absolute RD and offline hybrid prediction errors at step~120 using a shared error scale. This sample illustrates a compact tumor morphology in which the RD baseline preserves coarse location but underestimates heterogeneous cellularity, whereas the offline hybrid model better recovers internal density structure and reduces spatial error.
}
\label{fig:hybrid_qualitative_sample1}
\end{figure}

\begin{figure}[H]
\centering
\includegraphics[width=\textwidth]{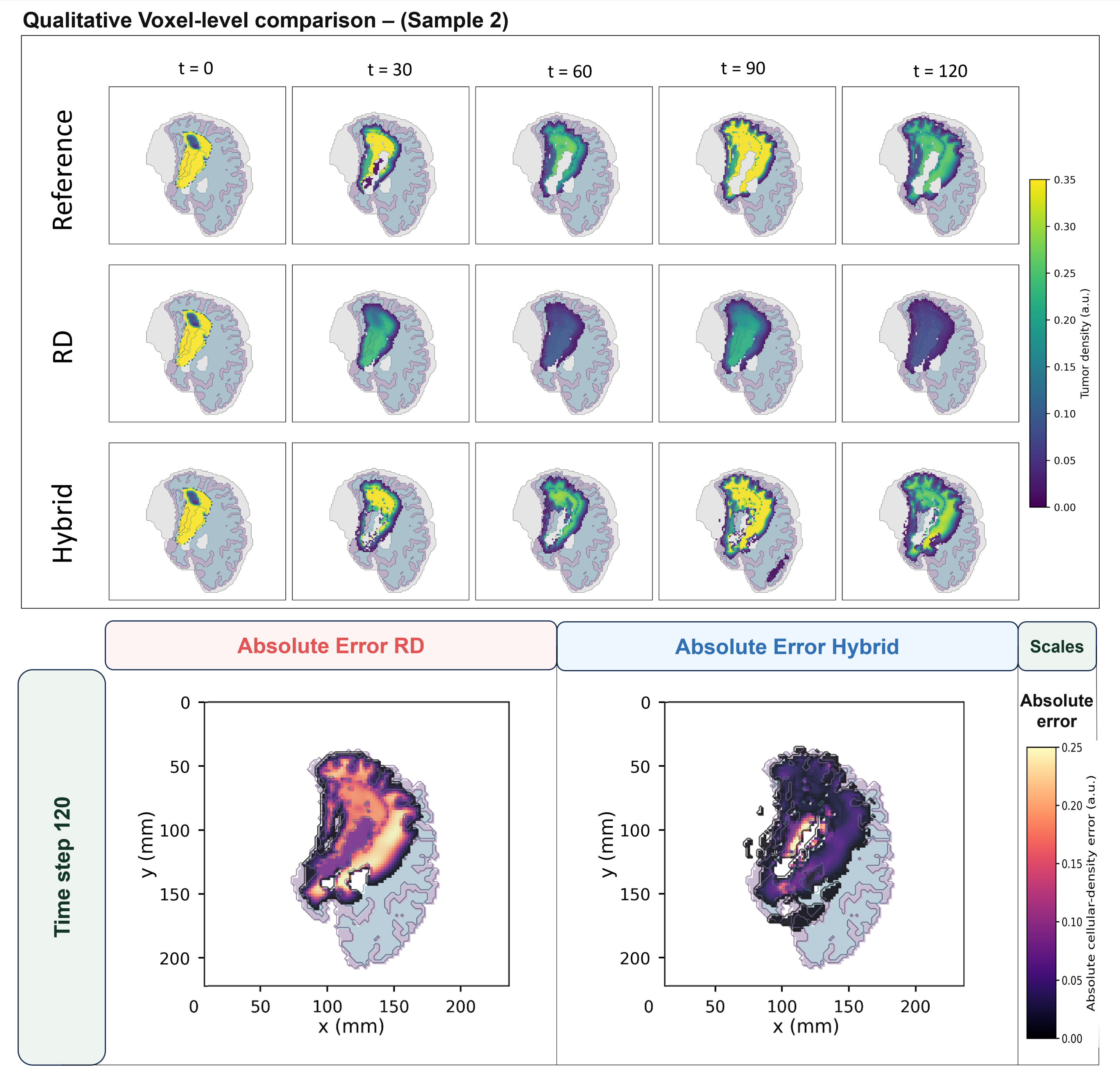}
\caption{
Qualitative voxel-level comparison for patient sample~2 in the dense
forecasting setting. The layout and reference-conditioned prediction
protocol follow Fig.~\ref{fig:hybrid_qualitative_sample1}; thus, the
displayed RD and hybrid outputs are one-step predictions rather than a
free-running recursive rollout from step~0. The tumor morphology and
spatial error pattern differ from patient sample~1. Together, the two
samples show that residual correction reduces RD model-form error across
different tumor shapes and cellularity distributions rather than only in
a single representative geometry.
}
\label{fig:hybrid_qualitative_sample2}
\end{figure}

Figures~\ref{fig:hybrid_qualitative_sample1} and~\ref{fig:hybrid_qualitative_sample2} show representative predictions for two patients with distinct tumor morphologies and density distributions.
In both cases, the RD model preserves the general tumor location but smooths internal spatial structure and underestimates higher-density regions.
The hybrid RD--residual model more closely reproduces the reference tumor field and substantially reduces voxel-wise error within the tumor-bearing region.

\begin{figure}[H]
\centering
\includegraphics[width=\textwidth]{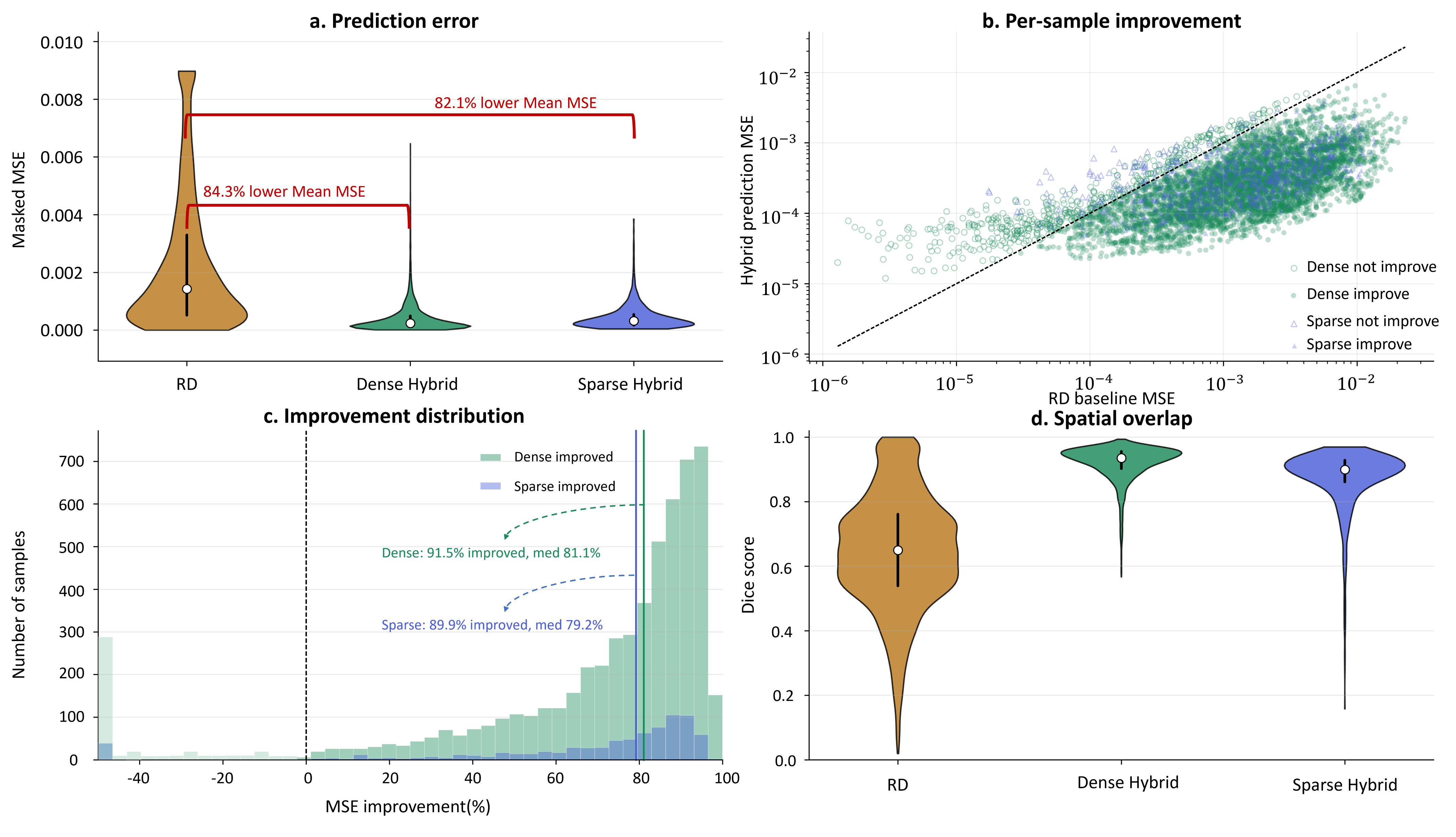}
\caption{
Quantitative evaluation of hybrid RD--residual forecasting performance. (a) Distribution of masked voxel-wise MSE for the RD baseline and hybrid models. (b) Per-sample comparison between RD and hybrid MSE; points below the dashed identity line indicate lower error after residual correction. (c) Distribution of percentage MSE improvement relative to the RD baseline, with positive values indicating improvement. (d) Distribution of Dice overlap scores. Dense and sparse sampling refer to different strides between sampled prediction-pair starting times.
}
\label{fig:hybrid_quantitative}
\end{figure}

\begin{table}[H]
\centering
\small
\setlength{\tabcolsep}{4pt}
\caption{
Hybrid forecasting performance on the held-out test cohort. Values are reported as mean $\pm$ standard deviation. Metrics compare model predictions against the controlled reference trajectory.
}
\label{tab:hybrid_forecasting_results}
\renewcommand{\arraystretch}{1.15}
\begin{tabular}{lcccc}
\toprule
Model & MSE & Dice & PSNR & Burden Error (mm$^3$) \\
\midrule
RD Baseline
&
$2.26 \times 10^{-3} \pm 2.72 \times 10^{-3}$
&
$0.641 \pm 0.189$
&
$29.85 \pm 6.47$
&
$12\,765 \pm 15\,340$
\\
Dense Hybrid
&
\textbf{$3.54 \times 10^{-4} \pm 5.17 \times 10^{-4}$}
&
\textbf{$0.920 \pm 0.057$}
&
\textbf{$36.99 \pm 4.47$}
&
\textbf{$2\,458 \pm 4\,729$}
\\
Sparse Hybrid
&
$4.05 \times 10^{-4} \pm 4.52 \times 10^{-4}$
&
$0.868 \pm 0.106$
&
$35.51 \pm 3.55$
&
$4\,846 \pm 3\,567$
\\
\bottomrule
\end{tabular}
\end{table}

Quantitative results are summarized in Fig.~\ref{fig:hybrid_quantitative} and Table~\ref{tab:hybrid_forecasting_results}.
In the dense reference-conditioned setting, masked voxel-wise MSE decreased from
$2.26\times10^{-3}$ for the RD baseline to
$3.54\times10^{-4}$ for the hybrid model, corresponding to an 84.3\% reduction.
Dice overlap increased from 0.641 to 0.920, an absolute increase of 0.279 and a relative increase of 43.5\%.
PSNR increased from 29.85 to 36.99~dB, while tumor-burden error decreased from
12,765 to 2,458~mm$^3$, corresponding to an 80.7\% reduction.

The sparse-sampling hybrid model also substantially improved reference-conditioned prediction.
MSE decreased to $4.05\times10^{-4}$, an 82.1\% reduction relative to the RD baseline.
Dice increased from 0.641 to 0.868, corresponding to a 35.4\% relative increase; PSNR increased by 5.66~dB; and tumor-burden error decreased by 62.0\%.
In the per-sample comparison in Fig.~\ref{fig:hybrid_quantitative}(b), most points fall below the identity line, indicating that residual correction reduces voxel-wise error for most evaluated samples. Positive MSE improvement is observed in 91.5\% of dense-sampling transitions and 89.9\% of sparse-sampling transitions.

These results demonstrate that the learned residual component effectively corrects the structured one-step prediction error of the RD model in the controlled validation setting.
Importantly, however, reference-conditioned accuracy does not directly establish stability during recursive forecasting, where each predicted state becomes the input to the next transition.
We therefore evaluated recursive performance separately.

\subsection{Patient-Specific Adaptation During Recursive Forecasting}
\label{subsec:online_adaptation}

% The next experiment evaluates the online-adaptation component of the proposed digital twin. This experiment addresses a different question from offline residual learning. The offline hybrid model improves one-step and short-horizon prediction when evaluated against reference states, but long-horizon deployment requires recursive rollout, where each predicted state becomes the input for the next prediction. The purpose of online adaptation is to reduce this patient-specific rollout drift by updating the residual component using recently available observations while keeping the RD backbone fixed.

We next evaluated whether patient-specific adaptation improves long-horizon recursive forecasting.
The RD, offline hybrid, and adapted hybrid models were initialized from the same step-0 tumor state and recursively propagated over 120 steps.
Unlike the reference-conditioned experiment above, each predicted state was used as the input to the subsequent transition.

\begin{figure}[H]
\centering
\includegraphics[width=1\textwidth]{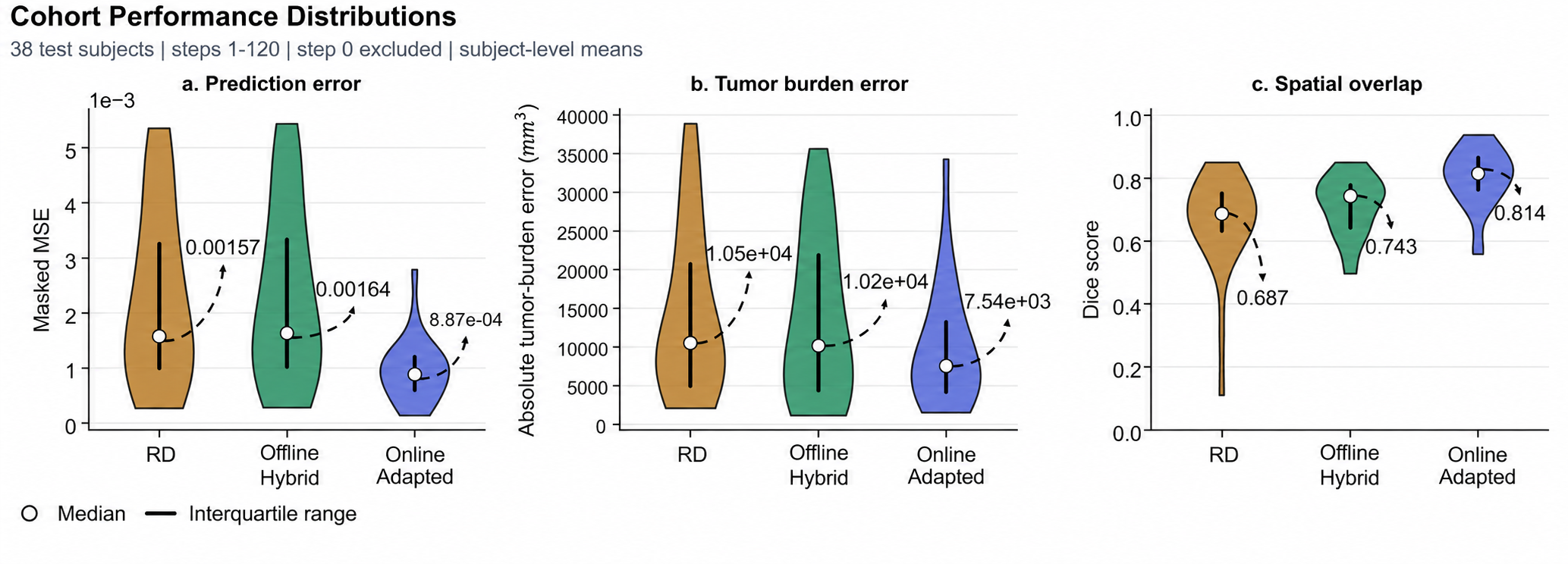}
\caption{
Cohort-level recursive forecasting performance for RD rollout, offline hybrid recursive rollout, and online-adapted recursive rollout across 38 held-out test patients. Metrics are computed as patient-level means over simulation steps~1--120, excluding the shared step-0 initialization. Panels show distributions of masked MSE, absolute tumor-burden error, and Dice overlap. White circles indicate medians, and black bars indicate interquartile ranges. Online adaptation lowers voxel-wise prediction error and tumor-burden error while improving spatial overlap relative to the offline hybrid model evaluated recursively without patient-specific updates.
}
\label{fig:online_cohort_distributions}
\end{figure}

% Figure~\ref{fig:online_cohort_distributions} summarizes recursive forecasting performance across 38 held-out test patients. Metrics are averaged over simulation steps~1--120, excluding the shared step-0 initialization. The offline hybrid model, when deployed recursively without online updates, improves Dice overlap relative to the RD baseline, increasing Dice from 0.687 to 0.743, but it does not reduce voxel-wise MSE relative to RD. This indicates that population-level residual correction improves tumor morphology but can still accumulate intensity and density-calibration errors during recursive deployment.

Figure~\ref{fig:online_cohort_distributions} summarizes performance across the 38 held-out test patients.
Metrics were first averaged over steps~1--120 for each patient, excluding the common step-0 initialization.
The offline hybrid model improved Dice relative to the RD model, from 0.687 to 0.743, but did not improve voxel-wise MSE, which changed from
$1.57\times10^{-3}$ to $1.64\times10^{-3}$.
This contrast indicates that improved one-step residual correction does not necessarily translate into improved recursive accuracy across all metrics.

\begin{table}[H]
\centering
\caption{
Cohort-level online adaptation performance. Values are cohort medians of patient-level time-averaged metrics across the 38 held-out patients. Each patient-level metric was averaged over simulation steps~1--120, excluding the shared step-0 initialization.
}
\label{tab:online_adaptation_cohort}
\begin{tabular}{lcccc}
\toprule
Model & MSE & Dice & PSNR & Burden Error (mm$^3$) \\
\midrule
RD
& $1.57 \times 10^{-3}$
& 0.687
& 28.0
& $1.05 \times 10^{4}$ \\
Offline Hybrid
& $1.64 \times 10^{-3}$
& 0.743
& 27.9
& $1.02 \times 10^{4}$ \\
Online Adapted
& $\textbf{8.87} \times \textbf{10}^{\textbf{-4}}$
& \textbf{0.814}
& \textbf{30.5}
& $\textbf{7.54} \times \textbf{10}^{\textbf{3}}$ \\
\bottomrule
\end{tabular}
\end{table}

% Online adaptation produces the strongest overall recursive performance. Compared with the offline hybrid model evaluated recursively without patient-specific updates, online adaptation reduces MSE from $1.64\times10^{-3}$ to $8.87\times10^{-4}$, increases Dice from 0.743 to 0.814, improves PSNR from 27.9 to 30.5 dB, and reduces volume error from $1.02\times10^4$ to $7.54\times10^3$ mm$^3$. These changes correspond to a 45.9\% reduction in MSE, a 9.6\% increase in Dice overlap, a 2.6 dB increase in PSNR, and a 26.1\% reduction in tumor-volume error relative to the offline hybrid recursive rollout. These results show that offline residual correction alone is not sufficient for stable patient-specific recursive forecasting. During rollout, local errors propagate forward and change the future input distribution, especially near tumor boundaries and high-density regions. Online adaptation addresses this issue by recalibrating the residual component using recent patient-specific observations while keeping the RD backbone fixed.

Under dense observations, patient-specific adaptation achieved the best recursive forecasting performance across the reported metrics. 
As shown in Table~\ref{tab:online_adaptation_cohort}, MSE decreased from
$1.64\times10^{-3}$ for the offline hybrid rollout to
$8.87\times10^{-4}$ after adaptation, corresponding to a 45.9\% reduction.
Dice increased from 0.743 to 0.814, a relative increase of 9.6\%.
PSNR increased from 27.9 to 30.5~dB, and tumor-burden error decreased from
$1.02\times10^4$ to $7.54\times10^3$~mm$^3$, corresponding to a 26.1\% reduction.

\begin{figure}
\centering
\includegraphics[width=1\textwidth]{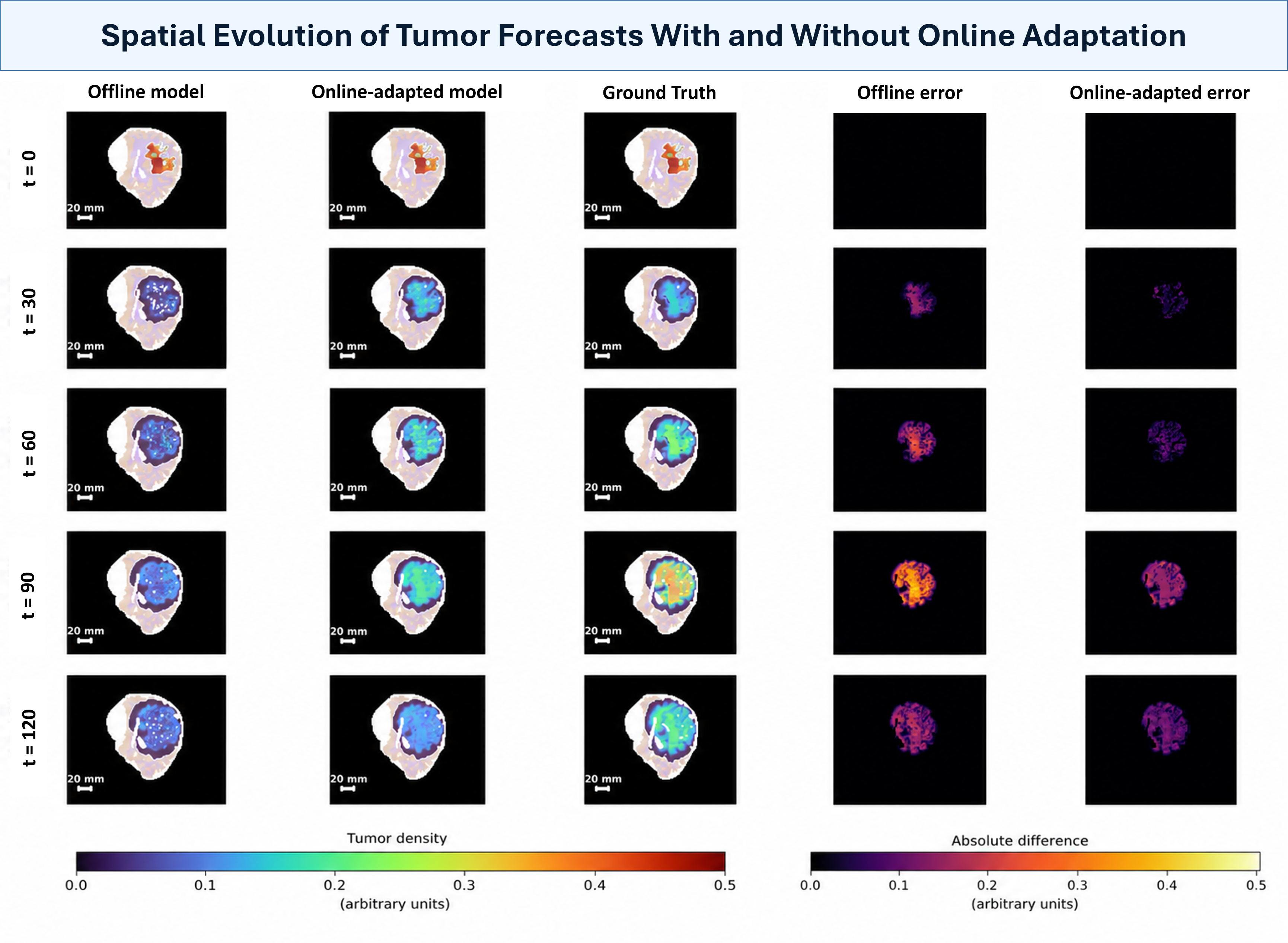}
\caption{
Spatial evolution of tumor forecasts with and without online adaptation for an illustrative patient sample. Columns show the offline hybrid recursive prediction, online-adapted recursive prediction, reference trajectory, offline hybrid prediction error, and online-adapted prediction error. Rows show selected rollout steps from initialization to simulation step~120. Online adaptation better preserves tumor density and spatial structure during long-horizon rollout, reducing localized density and boundary errors relative to the offline hybrid recursive forecast.
}
\label{fig:forecast_comp}
\end{figure}

% Figure~\ref{fig:forecast_comp} shows the effect of online adaptation for an illustrative patient trajectory. The offline hybrid model captures the approximate tumor location but progressively loses internal density structure during recursive rollout. In contrast, the online-adapted model remains more closely aligned with the reference trajectory, particularly at later time points when recursive drift becomes more pronounced. The reduction in error is concentrated within the tumor-bearing region, suggesting that online adaptation primarily corrects accumulated local mismatch rather than uniformly shifting the entire tumor field. 

% % Together, the cohort-level and patient-level results show that online adaptation improves recursive forecasting stability and provides a practical digital-twin updating mechanism in the controlled synthetic setting.

Figure~\ref{fig:forecast_comp} illustrates these differences for an individual patient.
The offline hybrid model retains the approximate tumor location but progressively loses internal density structure during recursive forecasting.
The adapted model remains more closely aligned with the reference trajectory, particularly at later time points, with reduced density and boundary errors within the tumor-bearing region.

% \delete{These findings distinguish population-level residual learning from patient-specific adaptation.}
% \delete{The former improves reference-conditioned prediction, whereas the latter reduces errors that accumulate when the hybrid model is deployed recursively.}

\subsection{Scenario-Based MPC Treatment Scheduling}
\label{subsec:treatment_planning_results}

% The fourth experiment evaluates the treatment-scheduling component of the framework. Because online adaptation achieved the strongest recursive forecasting performance, the online-adapted hybrid model was used inside the scenario-based robust MPC controller. The goal of this experiment is not to establish clinical treatment superiority, but to determine whether the proposed digital twin can be connected to a constrained receding-horizon planner and generate feasible adaptive treatment schedules. Both the fixed schedule and the MPC-selected schedule were replayed through the same controlled reference simulator before outcome metrics were computed.

% At each decision step, the controller used the current tumor estimate to evaluate candidate treatment sequences over a 10-step planning horizon, applied the first selected action, and repeated the optimization as the simulated trajectory advanced. This receding-horizon structure allows treatment actions to respond to updated tumor forecasts rather than committing to a fixed 120-step schedule at initialization. For outcome evaluation, the selected MPC schedule and the fixed comparator schedule were replayed through the controlled reference simulator, and tumor-burden endpoints were computed from the resulting simulator-generated trajectories.

The adapted hybrid DT was subsequently used as the predictive model within the MPC treatment-scheduling framework.
At each decision step, the controller evaluated candidate treatment sequences over a 10-step planning horizon, executed the first selected action, and replanned as the trajectory evolved.
The selected MPC schedules and the fixed comparator were then replayed through the same controlled reference simulator for outcome evaluation.

\begin{figure}[H]
\centering
\includegraphics[width=\textwidth]{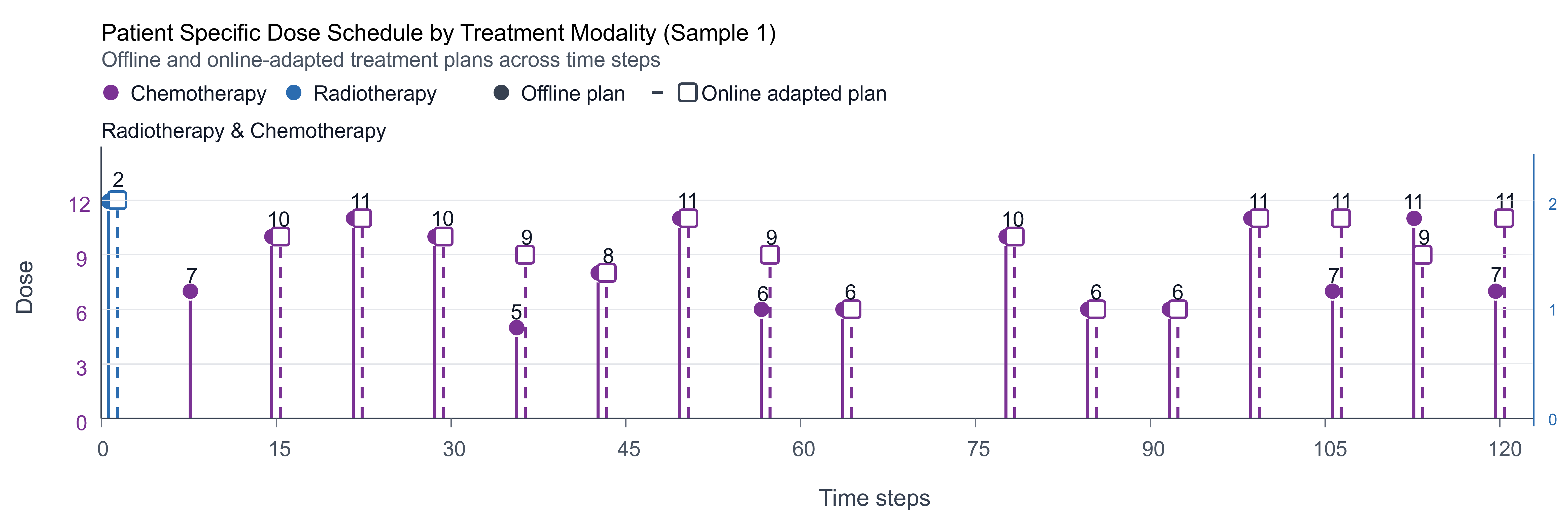}
\caption{
Patient-specific treatment schedules generated for an illustrative patient using offline and online-adapted scenario-based robust MPC planning. This figure illustrates controller behavior, whereas the cohort-level outcome comparison evaluates the fixed synthetic schedule against online-adapted MPC.
}
\label{fig:dose_schedule}
\end{figure}

% Figure~\ref{fig:dose_schedule} illustrates controller behavior for a representative patient. The generated schedules satisfy the modality and timing constraints and show sparse treatment actions over the 120-step simulation. The online-adapted controller modifies chemotherapy timing and intensity as the tumor forecast evolves, illustrating how the proposed framework shifts the digital twin from passive forecasting to decision-oriented treatment-scheduling simulation.

Figure~\ref{fig:dose_schedule} shows treatment schedules for an illustrative patient.
The MPC-generated schedules satisfy the prescribed modality and timing constraints and adjust chemotherapy timing and dose as the predicted tumor trajectory evolves.
This example illustrates the receding-horizon behavior of the controller rather than a cohort-level comparison of treatment efficacy.

\begin{table}[H]
\centering
\caption{
Treatment-scheduling comparison between the fixed schedule and online-adapted MPC across 38 held-out patients. Values are reported as median [interquartile range] after replaying each schedule through the same controlled reference simulator. Final burden measures terminal tumor burden at simulation step~120, while cumulative burden integrates tumor burden over the full rollout. Toxicity denotes the maximum value of the reduced chemotherapy-toxicity state; both strategies remain below the toxicity threshold of 5.0.
}
\label{tab:treatment_planning_online_mpc}
\resizebox{\textwidth}{!}{%
\begin{tabular}{lccccccc}
\toprule
Planning strategy & $n$ & Final burden & Cumulative burden & Total chemo & Total RT & Max toxicity & Toxicity violation steps \\
\midrule
Fixed schedule 
& 38 
& 4,610 [2,696, 5,599] 
& \textbf{677,145 [412,212, 816,161]} 
& 136 [136, 136] 
& 2 [2, 2] 
& \textbf{3.65 [3.65, 3.65]} 
& 0 [0, 0] \\
Online-adapted MPC 
& 38 
& \textbf{3,601 [2,085, 4,368]} 
& 859,409 [518,067, 1,036,383] 
& 134 [134, 134] 
& 2 [2, 2] 
& 4.98 [4.98, 4.98] 
& 0 [0, 0] \\
\bottomrule
\end{tabular}
}
\end{table}

% Table~\ref{tab:treatment_planning_online_mpc} summarizes the replay-based treatment-scheduling comparison. Online-adapted MPC reduces the median final tumor burden from 4,610 to 3,601, corresponding to a paired median reduction of 22.4\%. Both strategies use the same total radiotherapy exposure and similar total chemotherapy exposure, and both satisfy the simplified toxicity constraint for all patients. The comparison also reveals that the current controller improves terminal tumor control by using the available toxicity budget more aggressively. The median maximum toxicity is 4.98 under online-adapted MPC compared with 3.65 under the fixed schedule, while remaining below the toxicity threshold of 5.0. In addition, cumulative tumor burden is higher under online-adapted MPC than under the fixed schedule. This occurs because the implemented MPC objective minimizes worst-case terminal tumor burden over the planning horizon rather than cumulative burden over the full 120-step simulation.

Table~\ref{tab:treatment_planning_online_mpc} summarizes treatment-scheduling outcomes across the 38 held-out patients.
The median final tumor burden was 4,610~[2,696, 5,599] under the fixed schedule and
3,601~[2,085, 4,368] under adapted MPC.
Across paired patients, the median relative reduction in final tumor burden was 22.4\%.
The two strategies used similar total chemotherapy exposure
(136 versus 134 action units at the cohort median) and identical radiotherapy exposure
(2~Gy at the cohort median). MPC operated closer to the prescribed chemotherapy-toxicity limit.
Median maximum toxicity increased from 3.65 under the fixed schedule to 4.98 under MPC, while remaining below the threshold of 5.0, and no toxicity violations were observed.
In contrast to final burden, median cumulative tumor burden was higher under MPC:
859,409~[518,067, 1,036,383] compared with
677,145~[412,212, 816,161] under the fixed schedule.

\begin{figure}
\centering
\begin{subfigure}{0.49\textwidth}
    \centering
    \includegraphics[width=\textwidth]{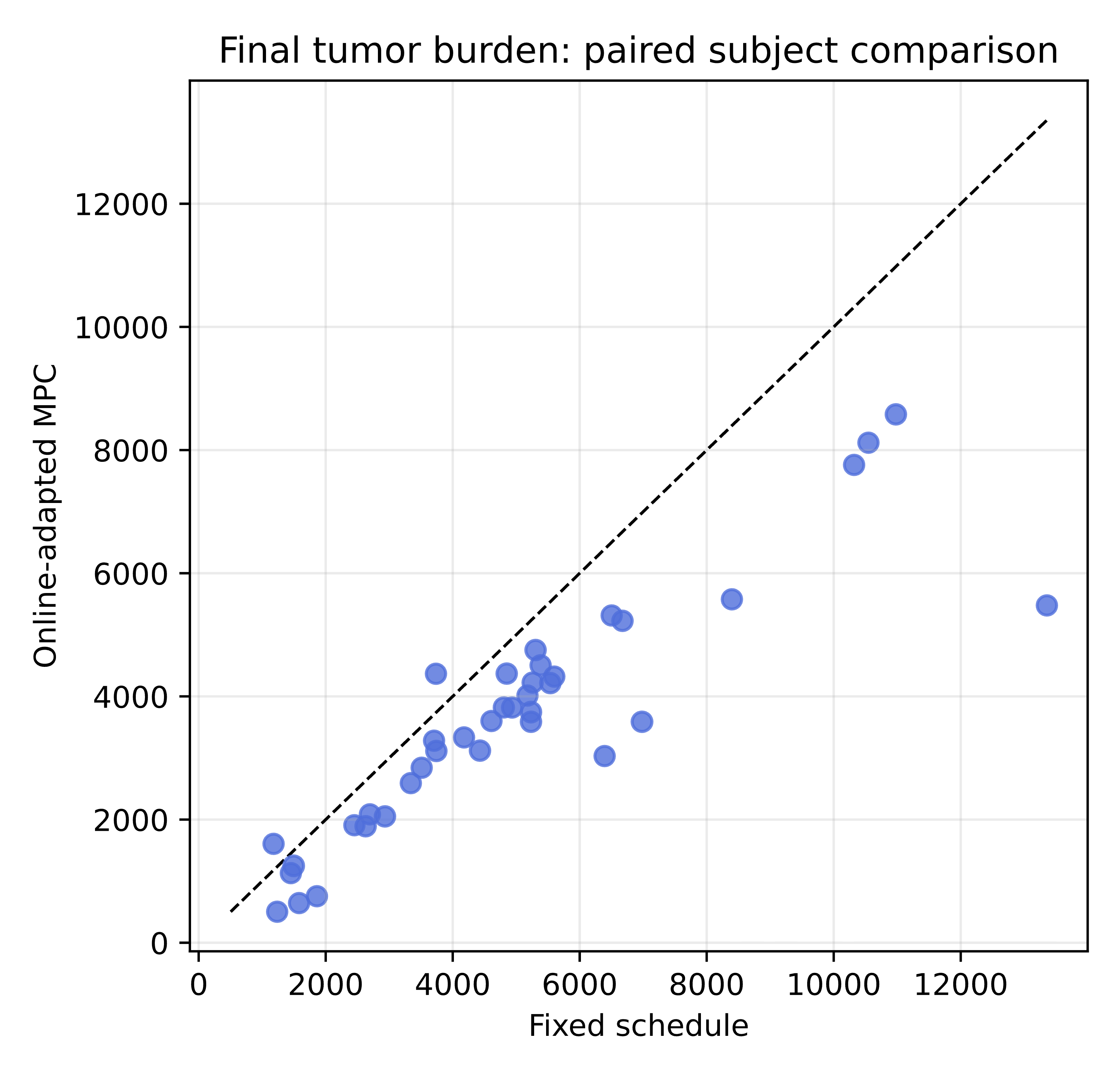}
    \caption{Final tumor burden.}
\end{subfigure}
\hfill
\begin{subfigure}{0.49\textwidth}
    \centering
    \includegraphics[width=\textwidth]{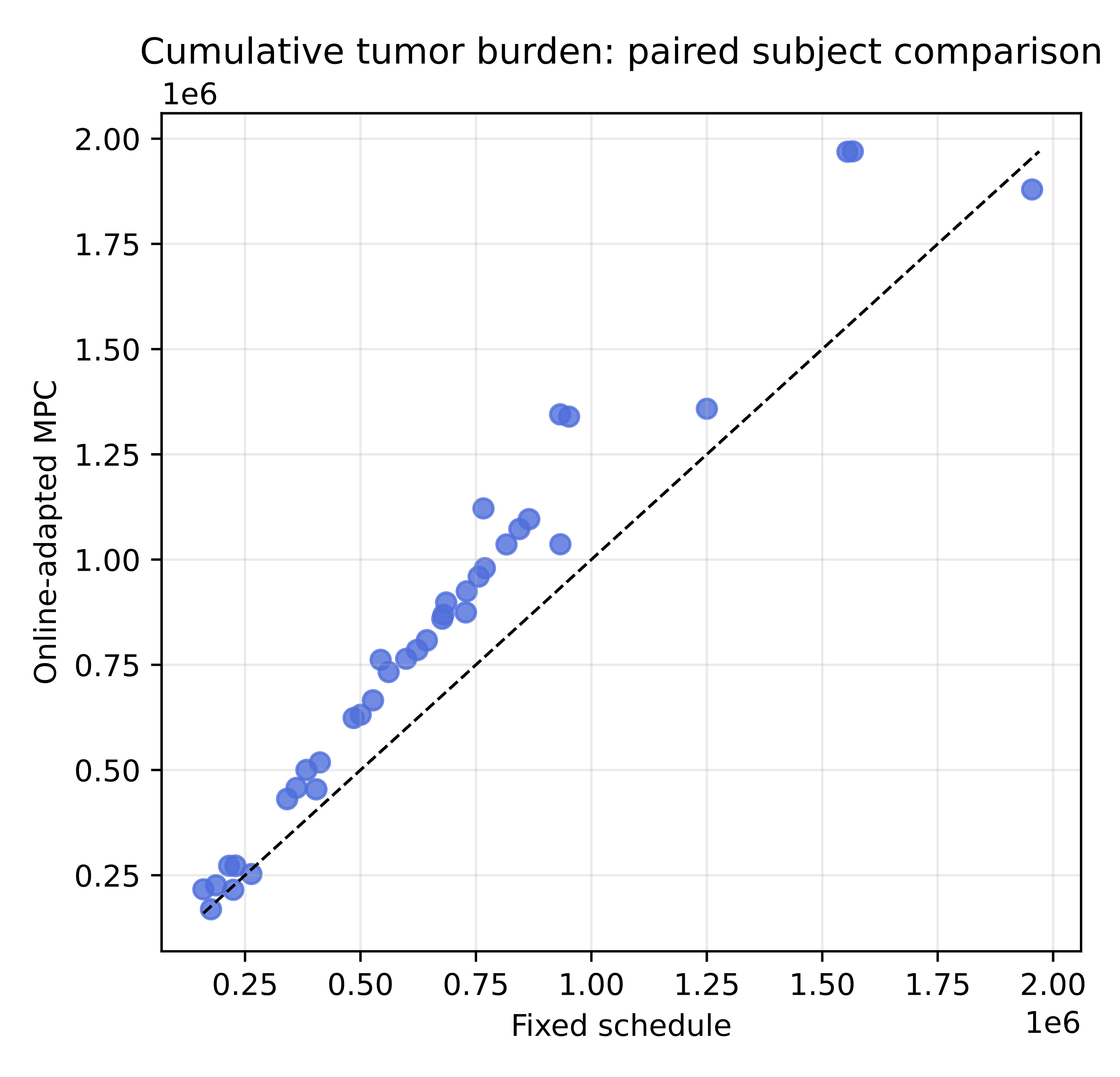}
    \caption{Cumulative tumor burden.}
\end{subfigure}
\caption{
Paired treatment-scheduling comparison between the fixed schedule and online-adapted MPC across 38 held-out patients after replaying both schedules through the same controlled reference simulator. Each point represents one patient, and the dashed line indicates equal tumor burden between strategies. Points below the dashed line favor online-adapted MPC, whereas points above the line favor the fixed schedule. Online-adapted MPC reduces final tumor burden for most patients but increases cumulative tumor burden, indicating that the present controller primarily optimizes terminal burden rather than total burden accumulated over time.
}
\label{fig:treatment_paired_comparison}
\end{figure}

% Figure~\ref{fig:treatment_paired_comparison} makes this tradeoff explicit. Online-adapted MPC reduces final tumor burden for most patients, but cumulative tumor burden is higher for most patients. Thus, the treatment-scheduling experiment demonstrates feasibility of the closed-loop forecasting-control pipeline, while also highlighting the importance of objective design. If the endpoint is terminal tumor burden, the current online-adapted MPC improves the simulated outcome relative to the fixed schedule. If the endpoint is cumulative tumor burden, the fixed schedule performs better in the current configuration. This distinction should guide future controller designs that balance terminal control, cumulative burden, dose exposure, and toxicity.

The paired comparison in Fig.~\ref{fig:treatment_paired_comparison} shows the same tradeoff at the patient level.
MPC reduces final tumor burden for most patients but increases cumulative burden for most patients.
This behavior is consistent with the controller objective, which prioritizes worst-case terminal tumor burden over the finite planning horizon rather than cumulative burden over the complete trajectory.
Thus, the scheduling experiment demonstrates that the adapted DT can be incorporated into constrained receding-horizon treatment planning, while also showing that the resulting treatment behavior depends strongly on the chosen optimization objective.

\subsection{Observation-Frequency Sensitivity Analysis}
\label{subsec:sensitivity_analysis}

% The final experiment examines forecasting sensitivity to the availability
% of reference observations. Dense and sparse configurations were evaluated
% using the corresponding RD, offline hybrid, and online-adapted models.
% To account for differences between the two configurations, performance
% changes are evaluated within each configuration relative to its corresponding RD baseline. This analysis therefore characterizes the end-to-end sensitivity of the forecasting framework to the observation schedule.

Finally, we examined the sensitivity of recursive forecasting to the frequency of available tumor-state observations.
Because the dense and sparse configurations differ in their sampling and adaptation protocols, comparisons were made within each configuration relative to its corresponding RD baseline. %rather than directly between dense and sparse settings.

\begin{table}[H]
\centering
\caption{
Observation-schedule sensitivity analysis across 38 held-out patients.
Values are cohort medians of patient-level metrics averaged over the selected evaluation time points for each configuration. All models within a configuration were evaluated at the same time points. Error reductions are relative to RD within each configuration; negative values indicate worse performance.}
\label{tab:observation_schedule_sensitivity}
\resizebox{\textwidth}{!}{%
\begin{tabular}{llccccc}
\toprule
Observation schedule & Model & Dice & MSE & Burden Error (mm$^3$) & MSE change vs. RD & Burden-error change vs. RD \\
\midrule
Dense
& RD 
& 0.687 
& $1.57\times10^{-3}$ 
& $1.05\times10^{4}$ 
& 0.0\% 
& 0.0\% \\
Dense
& Offline Hybrid 
& 0.743 
& $1.64\times10^{-3}$ 
& $1.02\times10^{4}$ 
& -3.9\% 
& 3.4\% \\
Dense
& Online Adapted 
& \textbf{0.814} 
& \textbf{$8.87\times10^{-4}$} 
& \textbf{$7.54\times10^{3}$} 
& \textbf{43.6\%} 
& \textbf{28.3\%} \\
\midrule
Sparse 
& RD 
& 0.759 
& $2.31\times10^{-3}$ 
& $1.11\times10^{4}$ 
& 0.0\% 
& 0.0\% \\
Sparse
& Offline Hybrid 
& 0.741 
& \textbf{$2.26\times10^{-3}$} 
& $1.27\times10^{4}$ 
& 2.2\% 
& -14.6\% \\
Sparse 
& Online Adapted 
& \textbf{0.880} 
& $2.66\times10^{-3}$ 
& $1.24\times10^{4}$ 
& -15.2\% 
& -11.8\% \\
\bottomrule
\end{tabular}
}
\end{table}

% Under dense observations, online adaptation improves all reported metrics relative to both RD rollout and offline hybrid recursive rollout. Dice increases to 0.814, MSE decreases by 43.6\% relative to RD, and volume error decreases by 28.3\%. This confirms that, when sufficiently frequent observations are available, patient-specific updating can improve spatial overlap, voxel-level accuracy, and tumor-burden estimation during recursive rollout.

Under the dense-observation configuration, patient-specific adaptation produced the best performance across the reported metrics.
Dice increased from 0.687 for RD to 0.814 after adaptation, while MSE decreased from
$1.57\times10^{-3}$ to $8.87\times10^{-4}$ and tumor-burden error decreased from
$1.05\times10^4$ to $7.54\times10^3$~mm$^3$.
These results correspond to approximately 43.6\% lower MSE and 28.3\% lower burden error relative to the corresponding RD baseline.

% The sparse-observation setting shows different behavior. Online adaptation achieves the highest Dice score among the sparse-observation models, increasing Dice to 0.880, but its MSE and volume error are worse than the sparse-observation RD baseline. This discrepancy indicates that Dice alone does not capture density calibration. The sparse-observation online-adapted model can preserve approximate tumor support while misestimating cellularity magnitude inside that support. For treatment scheduling, this distinction is important because the controller depends on tumor burden and cellularity estimates, not only thresholded spatial overlap.

The reduced-frequency configuration showed a different pattern.
The adapted model achieved the highest Dice score, increasing from 0.759 for RD to 0.880.
However, MSE increased from
$2.31\times10^{-3}$ to $2.66\times10^{-3}$, and tumor-burden error increased from
$1.11\times10^4$ to $1.24\times10^4$~mm$^3$.
Thus, adaptation improved tumor-support overlap but did not improve voxel-level density or tumor-burden error under the reduced-frequency observation setting.

\begin{figure}[H]
\centering
\includegraphics[width=\textwidth]{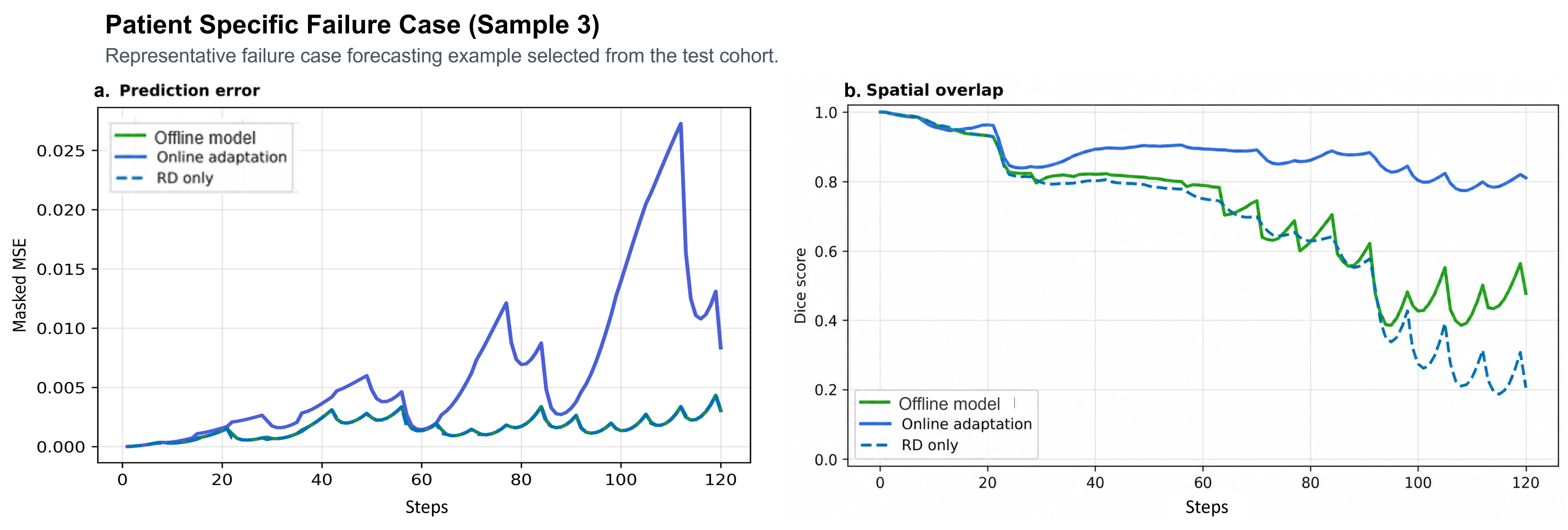}
\caption{
Representative patient-specific failure case for online adaptation. Panel (a) shows masked voxel-wise MSE over the 120-step rollout for RD-only, offline hybrid recursive forecasting, and online-adapted recursive forecasting. Panel (b) shows the corresponding Dice overlap. In this case, online adaptation maintains higher spatial overlap than RD and offline hybrid forecasting for much of the rollout, but it produces large transient MSE increases during later time points. The example illustrates that online adaptation can improve tumor-support alignment while failing to maintain calibrated voxel-level density.
}
\label{fig:online_adaptation_failure_case}
\end{figure}

% Figure~\ref{fig:online_adaptation_failure_case} illustrates this failure mode. Online adaptation maintains relatively high spatial overlap, especially compared with RD-only forecasting near the end of the rollout, but it produces large MSE spikes during later simulation steps. The failure is therefore not loss of tumor localization; it is a mismatch between tumor-support accuracy and voxel-level density calibration.

Figure~\ref{fig:online_adaptation_failure_case} illustrates this behavior for an individual patient.
The adapted model maintains relatively high Dice overlap during much of the rollout but exhibits large transient increases in MSE at later time points.
The failure therefore reflects inaccurate voxel-level density despite preservation of approximate tumor support.

% This sensitivity analysis supports two conclusions. First, the dense-observation online-adapted model provides the strongest overall balance of MSE, Dice, PSNR, and volume-error performance in the controlled experiments, which motivates its use in the treatment-scheduling simulation. Second, online adaptation is not automatically robust to sparse observations. The sparse-observation experiment reused the dense-observation adaptation hyperparameters, including learning rate, number of update steps, buffer size, and rollout horizon. With fewer patient-specific transitions, adaptation requires separate tuning and uncertainty-aware safeguards. Future work should therefore develop sparse-observation adaptation strategies and evaluate treatment scheduling under clinically realistic imaging intervals.

% \delete{Overall, the sensitivity analysis shows that the benefit of patient-specific adaptation depends on observation availability. Frequent observations improved both spatial overlap and density-based accuracy, whereas reduced-frequency adaptation improved Dice but degraded MSE and volume error in the present configuration.} 
Overall, adaptation improved both spatial overlap and density-based accuracy within the dense configuration. Within the sparse configuration, adaptation improved Dice but degraded MSE and burden error relative to the corresponding RD baseline. Because the configurations used different evaluation time points, these results do not isolate the effect of observation frequency alone. 
Because the reduced-frequency experiment reused the adaptation hyperparameters selected for the dense-observation setting, these results also indicate the need for adaptation strategies specifically designed for sparse longitudinal observations.

\section{Discussion}

% \delete{This study develops an AI-augmented adaptive DT framework that integrates patient-specific initialization, mechanistic tumor modeling, learned model-form correction, patient-specific adaptation, and MPC-based treatment scheduling. The methodological contribution is the integration of hybrid forecasting, patient-specific adaptation, and treatment scheduling through a common treatment-conditioned tumor-evolution model. The RD model provides an interpretable dynamical backbone, while the 3D residual network corrects structured prediction errors not captured by the prescribed equations. Patient-specific adaptation is then used to reduce drift during recursive forecasting, and the adapted DT serves as the predictive model for evaluating constrained future treatment actions through MPC.}

The baseline RD model captured the coarse tumor location and overall temporal pattern but accumulated errors in tumor density, morphology, and burden during long recursive rollouts. In the present controlled setting, this behavior arises primarily from representing spatially heterogeneous proliferation with a spatially constant proliferation parameter. The hybrid RD--residual model substantially reduced this discrepancy under reference-conditioned forecasting, showing that the learned component can compensate for structured model-form error while retaining the RD model as the explicit tumor-evolution backbone. The residual term should therefore be interpreted as a learned correction to an incomplete mechanistic model rather than as an independently identified biological mechanism. These findings complement previous imaging-informed mechanistic tumor models that use longitudinal measurements for patient-specific calibration and forecasting \cite{hormuth2015predicting,wu2025mri}, and are related to hybrid mechanistic--learning approaches that combine dynamical models with data-driven components \cite{mascheroni2021improving,martens2022deep}. The present study extends this direction by examining residual correction under recursive forecasting, patient-specific adaptation, and subsequent integration with treatment scheduling. Because the longitudinal trajectories are simulator generated, these comparisons establish methodological feasibility rather than clinical superiority. 
% \delete{The recursive experiments highlight an important distinction between reference-conditioned prediction and long-horizon DT deployment. The offline hybrid model achieved strong one-step correction but did not maintain the same improvement across all metrics when recursively driven by its own predictions. Errors introduced during rollout alter the future input distribution and can progressively shift the predicted trajectory away from the evolving patient state. Patient-specific adaptation reduced this drift by recalibrating the residual component using newly available observations while leaving the RD backbone unchanged. The resulting improvement indicates that population-level residual learning and patient-specific adaptation play complementary roles: the former corrects systematic model-form error across patients, whereas the latter adjusts the learned correction to the trajectory encountered during deployment. More broadly, these results show that strong one-step prediction accuracy does not necessarily imply stable long-horizon forecasting.} 

% The offline hybrid model improved one-step prediction but did not consistently improve recursive forecasts. Patient-specific adaptation reduced this drift while keeping the RD backbone fixed. These results distinguish population-level residual learning from patient-specific updating during deployment.

The offline hybrid model improved reference-conditioned one-step prediction, but these gains did not consistently carry over to recursive forecasting. Repeated use of predicted states can propagate errors and shift the input distribution. With frequent observations, patient-specific adaptation reduced this drift by updating the residual network while preserving the RD backbone. These findings highlight the complementary roles of population-level model-form correction and patient-specific updating during deployment.

The observation-schedule comparison shows different adaptation outcomes within the dense and sparse evaluation configurations. Because evaluation sampling also differed, the cross-configuration differences cannot be attributed solely to observation frequency. With frequent synthetic observations, adaptation improved both spatial overlap and density-based accuracy. Under the reduced-frequency setting, however, adaptation produced higher Dice overlap while worsening MSE and tumor-burden error relative to the corresponding RD baseline. This difference shows that preservation of tumor support does not necessarily imply accurate calibration of tumor density or burden. The observation schedule should therefore be considered part of the DT design rather than only an experimental parameter. Translation to realistic longitudinal settings will require adaptation strategies that account for the interval between observations, uncertainty in the observed tumor state, and the risk of overcorrection when only limited patient-specific information is available.

Coupling the adapted DT with MPC demonstrates how predictive modeling can be extended from forecasting to decision-oriented treatment evaluation. Within the controlled setting, MPC generated feasible chemotherapy and radiotherapy action schedules subject to timing, dose, toxicity, and bounded state-perturbation constraints. The resulting schedules reduced terminal tumor burden relative to the fixed comparator for most patients, but they also operated closer to the toxicity limit and produced higher cumulative tumor burden. This tradeoff is consistent with the implemented MPC objective, which prioritizes worst-case terminal burden over the finite planning horizon. The result emphasizes that improved prediction alone does not determine a desirable treatment policy. Decision quality also depends on the objective function, constraints, planning horizon, and the clinical quantities prioritized during optimization. Future controller designs should therefore consider multi-objective formulations that balance terminal tumor control, cumulative burden, treatment exposure, and toxicity.

% \delete{Several limitations remain. First, longitudinal tumor evolution and treatment response were generated computationally rather than observed clinically. The controlled simulator was designed to isolate model-form correction, recursive adaptation, and treatment-scheduling behavior, but it does not capture the full complexity of brain tumor progression. Important processes such as treatment resistance, edema, necrosis, evolving microenvironmental effects, imaging noise, and segmentation uncertainty are not represented. Second, the primary adaptation results rely on more frequent observations than would typically be available in clinical practice, and the reduced-frequency analysis shows that the current adaptation procedure is not automatically robust when observations become less frequent. Third, the MPC treatment model is intentionally simplified and is intended to evaluate computational scheduling behavior rather than reproduce clinical chemotherapy or radiotherapy protocols. Finally, the uncertainty treatment used in the current controller is limited to bounded perturbations of the estimated tumor state and does not represent broader uncertainty in model parameters, treatment response, or future observations.}

% Validation used synthetic trajectories rather than observed longitudinal outcomes. The model omits treatment resistance and imaging uncertainty, and its treatment dynamics are simplified. Adaptation under sparse observations remains limited, while MPC uncertainty covers only bounded state perturbations.

Validation used patient-informed synthetic trajectories rather than observed longitudinal outcomes, enabling controlled evaluation without capturing the full complexity of tumor progression and treatment response. The model omits treatment resistance and imaging uncertainty, while chemotherapy, radiotherapy, and toxicity are represented using simplified dynamics. Adaptation under sparse observations remains limited, particularly in maintaining accurate tumor-density and burden estimates despite improved spatial overlap. Finally, MPC uncertainty is restricted to bounded perturbations of the estimated tumor state and does not account for uncertainty in model parameters, treatment response, or future observations.

Despite these limitations, the patient-informed initialization preserves individualized anatomy and tumor structure, while the known longitudinal trajectories provide a controlled environment in which forecasting error, adaptation behavior, and treatment decisions can be evaluated separately. A natural next step is validation in longitudinal animal tumor models, where repeated imaging, controlled treatment delivery, biological response, and terminal histopathology could provide a more realistic yet experimentally controlled assessment. Subsequent clinical studies will require real longitudinal imaging, delivered treatment records, sparse and irregular observations, uncertainty-aware state estimation and adaptation, and treatment objectives aligned with clinically meaningful outcomes.

% \section{Conclusion}

% We developed an adaptive tumor DT framework that combines an RD model with residual learning, patient-specific updating, and MPC-based treatment scheduling. In the controlled patient-data-informed synthetic experiments, the RD model reproduced the overall pattern of tumor evolution but did not fully capture spatial heterogeneity over long prediction horizons. Adding the residual model improved the forecasts, and updating this component when new observations became available further reduced drift during recursive prediction. When used within MPC, the updated DT supported comparison of constrained chemotherapy and radiotherapy policies and improved terminal tumor control
% under the selected objective, while revealing tradeoffs in cumulative tumor burden and toxicity. These findings demonstrate the technical feasibility of combining mechanistic and learning-based models for adaptive tumor prediction and treatment-scheduling studies. The current results are based on controlled synthetic trajectories and should not be interpreted as clinical validation. Longitudinal animal studies would provide a practical next step for evaluating the framework with repeated imaging, controlled treatment delivery, and biological response measurements before clinical validation.

\section{Conclusion}

We developed an AI-augmented adaptive digital twin framework that integrates mechanistic reaction--diffusion modeling, learned residual correction, patient-specific adaptation, and MPC-based treatment scheduling. In the patient-informed synthetic validation setting, residual learning substantially improved tumor-state prediction beyond the RD model, while patient-specific adaptation further reduced error accumulation during recursive forecasting. When coupled with MPC, the adapted DT enabled constrained treatment-scheduling evaluation and reduced terminal tumor burden under the selected objective, while revealing tradeoffs in cumulative burden and toxicity. 
These findings establish the feasibility of the integrated prediction–adaptation–scheduling workflow in a controlled methodological setting. Longitudinal animal studies with repeated imaging and controlled treatment delivery provide a practical next step for evaluating the framework under more realistic biological and treatment conditions before clinical studies.

% \section*{Declaration of competing interest} 
% The authors declare that they have no known competing financial interests or personal relationships that could have appeared to influence the work reported in this paper.

% \section*{Declaration of generative AI and AI-assisted technologies in the manuscript preparation process} 
% During the preparation of this work the author(s) used ChatGPT in order to improve language and readability. After using this tool/service, the author(s) reviewed and edited the content as needed and take(s) full responsibility for the content of the publication.

% \section*{Acknowledgment}
% X. Li's work was partially funded by the Division of Mathematical Sciences, National Science Foundation with the award numbers 2410678. 

\section*{Acknowledgements and Declarations}

\subsection*{Ethical Approval}

No new human or animal participants were enrolled in this study. The
computational experiments used publicly available, de-identified imaging
data from the UPENN-GBM collection and simulator-generated longitudinal
trajectories. 

\subsection*{Funding}

Xianqi Li's work was partially supported by the Division of Mathematical
Sciences of the National Science Foundation under Award No.~2410678.

\subsection*{Competing Interests}

The authors declare that they have no known competing financial interests
or personal relationships that could have influenced the work reported
in this manuscript.

\subsection*{Declaration of Generative Artificial Intelligence and
Artificial Intelligence-Assisted Technologies}

During the preparation of this manuscript, the authors used ChatGPT to
improve language and readability. The authors subsequently reviewed and
edited the manuscript and take full responsibility for its content.

\bibliographystyle{elsarticle-num}
\bibliography{reference}
\end{document}